\def\paperID{12686}
\def\confName{CVPR}
\def\confYear{2024}
\def\paperTitle{
	Accurate Training Data
	for Occupancy Map Prediction\\in Automated Driving
	Using Evidence Theory
}

\def\authorBlock{
Jonas Kälble\footnotemark[2] \qquad
Sascha Wirges\footnotemark[2] \qquad
Maxim Tatarchenko\footnotemark[2] \qquad
Eddy Ilg\footnotemark[3] \\
\footnotemark[2] Bosch Center for Artificial Intelligence, Germany \\
\footnotemark[3] Saarland University, Germany \\
{\tt\small \{firstname.lastname\}@de.bosch.com} \qquad
{\tt\small ilg@cs.uni-saarland.de}
}

\newif\ifreview{}
\newif\ifarxiv{}
\newif\ifcamera{}
\newif\ifrebuttal{}

\PassOptionsToPackage{%
    hidelinks,
    pagebackref,
    breaklinks,
    citecolor=cvprblue,
    pdftitle={\paperTitle},
    }{hyperref}
\documentclass[10pt,twocolumn,letterpaper]{article}

\usepackage[pagenumbers]{cvpr}

\usepackage[T1]{fontenc}
\usepackage[utf8]{inputenc}
\usepackage[ngerman,main=english]{babel}
\usepackage{lmodern}
\usepackage{microtype}

\usepackage{graphicx}	
\usepackage{amsfonts, amsmath, amssymb}
\usepackage{booktabs}

\usepackage{adjustbox}
\usepackage{booktabs}
\usepackage{microtype}
\usepackage{epsfig}
\usepackage[table,xcdraw,dvipsnames]{xcolor}
\usepackage{caption}
\usepackage{phfparen}
\usepackage{mathdots}
\usepackage{mathtools}
\usepackage{float}
\usepackage{placeins}
\usepackage{color, colortbl}
\usepackage{stfloats}
\usepackage{enumitem}
\usepackage{siunitx}
\usepackage{tabularx}
\usepackage{multirow}
\usepackage{mfirstuc}
\usepackage{subcaption}
\usepackage{stackengine}
\usepackage[hang,flushmargin]{footmisc}
\usepackage{acro}
\usepackage{pdfcomment}
\usepackage{pgfplots, pgfplotstable}
\usepackage{tikz}
\usepackage{times}
\usepackage{titlecaps}
\usepackage{xcolor}
\usepackage{xspace}
\usepackage{xstring}
\usepackage{xr-hyper}

\usepackage{mwe} %

\usepackage[capitalize]{cleveref} %

\ifarxiv{} \fi

\newcommand{\R}[1]{{%
            \textbf{%
                \ifstrequal{#1}{1}{\textcolor{red}{R#1}}{%
                    \ifstrequal{#1}{2}{\textcolor{blue}{R#1}}{%
                        \ifstrequal{#1}{3}{\textcolor{magenta}{R#1}}{%
                            \ifstrequal{#1}{4}{\textcolor{teal}{R#1}}{%
                                \textcolor{cyan}{R#1}%
                            }}}}%
            }%
        }}

\makeatletter
\newcommand*{\addFileDependency}[1]{
    \typeout{(#1)}
    \@addtofilelist{#1}
    \IfFileExists{#1}{}{\typeout{No file #1.}}
}

\makeatother

\newcommand{\bba}[1]{\func{m}`*(#1)}

\newcommand{\func}[1]{\mathrm{#1}}

\renewcommand{\vec}[1]{\mathbf{#1}}

\definecolor{cvprblue}{rgb}{0.21,0.49,0.74}

\newcommand{\rankone}[1]{\textbf{\underline{#1}}}
\newcommand{\ranktwo}[1]{\textbf{#1}}

\newrobustcmd*{\trigreen}{\tikz{\filldraw[draw=black!30!green,fill=black!30!green] (0,0) -- (1mm,0) -- (0.5mm,1mm);}}
\newrobustcmd*{\trired}{\tikz{\filldraw[draw=black!30!red,fill=black!30!red] (0,0) -- (1mm,0) -- (0.5mm,1mm);}}
\newrobustcmd*{\higherbetter}{\tikz{\stackanchor[1pt]{\trigreen}{\rotatebox{180}{\trired}}}}
\newrobustcmd*{\lowerbetter}{\tikz{\stackanchor[1pt]{\trired}{\rotatebox{180}{\trigreen}}}}

\acsetup{%
	format/list=\titlecap,
	make-links,
	pdfcomments/use,
	single,
	use-id-as-short,
}

\DeclareAcronym{AV}{long=automated vehicle, pdfcomment=Automated Driving}
\DeclareAcronym{BBA}{long=basic belief assignment, pdfcomment=Basic Belief Assignment}
\DeclareAcronym{BCE}{long=binary cross-entropy, pdfcomment=Binary Cross-Entropy}
\DeclareAcronym{CE}{long=cross-entropy, pdfcomment=Cross-Entropy}
\DeclareAcronym{DST}{long=Dempster-Shafer theory, pdfcomment=Dempster-Shafer Theory}
\DeclareAcronym{ET}{long=evidence theory, pdfcomment=Evidence Theory}
\DeclareAcronym{FOV}{long=field of view, pdfcomment=Field of View}
\DeclareAcronym{GT}{long=ground-truth, pdfcomment=Ground-Truth}
\DeclareAcronym{LIDAR}{short=LiDAR, long=light detection and ranging, pdfcomment=Light Detection and Ranging}
\DeclareAcronym{MAE}{long=mean absolute error, pdfcomment=Mean Absolute Error}
\DeclareAcronym{MSE}{long=mean squared error, pdfcomment=Mean Squared Error}
\DeclareAcronym{RMSE}{long=root mean squared error, pdfcomment=Root Mean Squared Error}
\DeclareAcronym{SotA}{short=SOTA, long=state-of-the-art, pdfcomment=State-of-the-Art}

\crefname{section}{Sec.}{Secs.}
\crefname{table}{Tab.}{Tabs.}
\crefname{figure}{Fig.}{Figs.}

\usepgfplotslibrary{%
	colorbrewer,
	groupplots,
}
\pgfplotsset{%
	bar cycle list/.style={%
			cycle list name=Set1,
			every axis plot/.append style={fill, thin},
		},
	compat=1.18,
	colormap/Spectral,
	colormap={reverse Spectral}{%
			indices of colormap={%
					\pgfplotscolormaplastindexof{Spectral}, ..., 0 of Spectral}
		},
	cycle list/Set1,
	cycle list name=Set1,
	enlarge x limits=0.05,
	enlarge y limits=0.05,
	every axis/.append style={semithick, font=\small},
	every axis plot/.append style={thick, font=\small},
	grid style={%
			opacity=0.5,
		},
	group/xticklabels at=edge bottom,
	group/yticklabels at=edge left,
	group/xlabels at=edge bottom,
	group/ylabels at=edge left,
	scaled x ticks=false,
	scaled y ticks=false,
	tick label style={font=\footnotesize},
	title style={%
			anchor=north east,
			at={(0.99, 0.95)},
			fill opacity=0.5,
			fill=white,
			font=\small,
			text opacity=1,
		},
	width=0.9\columnwidth,
}
\pgfplotstableset{%
	col sep=comma,
	precision=3,
}

\robustify\bfseries
\begin{document}
\title{\paperTitle}
\author{\authorBlock}
\maketitle

\begin{abstract}
	Automated driving fundamentally requires knowledge about the surrounding geometry of the scene.
	Modern approaches use only captured images to predict occupancy maps that represent the geometry.
	Training these approaches requires accurate data that may be acquired with the help of \acs{LIDAR} scanners.
	We show that the techniques used for current benchmarks and training datasets to convert \acs{LIDAR} scans into occupancy grid maps yield very low quality,
	and subsequently present a novel approach using evidence theory that yields more accurate reconstructions.
	We demonstrate that these are superior by a large margin, both qualitatively and quantitatively, and that we additionally obtain meaningful uncertainty estimates.
	When converting the occupancy maps back to depth estimates and comparing them with the raw \acs{LIDAR} measurements, our method yields a \acs{MAE} improvement of 30\% to 52\% on nuScenes and 53\% on Waymo over other occupancy ground-truth data.
	Finally, we use the improved occupancy maps to train a \acl{SotA} occupancy prediction method and demonstrate that it improves the \acs{MAE} by 25\% on nuScenes.
\end{abstract}

\section{Introduction}\label{sec:intro}

The safe operation of automated vehicles requires accurate environment models to reason about geometry and semantics.
The most popular geometry model today is an occupancy grid map that tessellates the space into a regular grid of volume elements called voxels~\cite{foley1996computer}.
Such occupancy grid maps allow for compute-efficient free-space estimation (e.g., via ray casting) and data storage via Octrees~\cite{wang2017ocnn, tatarchenko2017octree}.

For several years, automated driving has mainly relied on \ac{LIDAR} due to its low and distance-independent measurement error.
However, recent approaches attempt to reconstruct the 3D environment from images~\cite{tong2023scene, zhang2023occformer, huang2021bevdet, li2023fbocc} as they argue \ac{LIDAR} is sparse, expensive, and redundant\footnote{https://techcrunch.com/2019/04/22/anyone-relying-on-lidar-is-doomed-elon-musk-says/}.

\begin{figure}[t]
	\centering
	\begin{adjustbox}{max width=\columnwidth}
		\begin{tikzpicture}
			\begin{groupplot}[
					axis equal image,
					colormap/RdBu,
					enlargelimits=0,
					group style={%
							group size=2 by 2,
							horizontal sep=2mm,
							vertical sep=2mm,
						},
					point meta max=3.75,
					point meta min=-3.75,
					scale only axis,
					scatter src=explicit,
					title style={%
							at={(0.995, 0.95)},
						},
					width=0.45\columnwidth,
					xmax=1920,
					xmin=0,
					ymax=0,
					ymin=-1080,
					xticklabels=\empty,
					yticklabels=\empty,
				]
				\nextgroupplot[
					title=Occ3D \cite{tian2023occ3d},
					ylabel=GT,
				]
				\addplot graphics[xmax=1920, xmin=0, ymax=0, ymin=-1080]{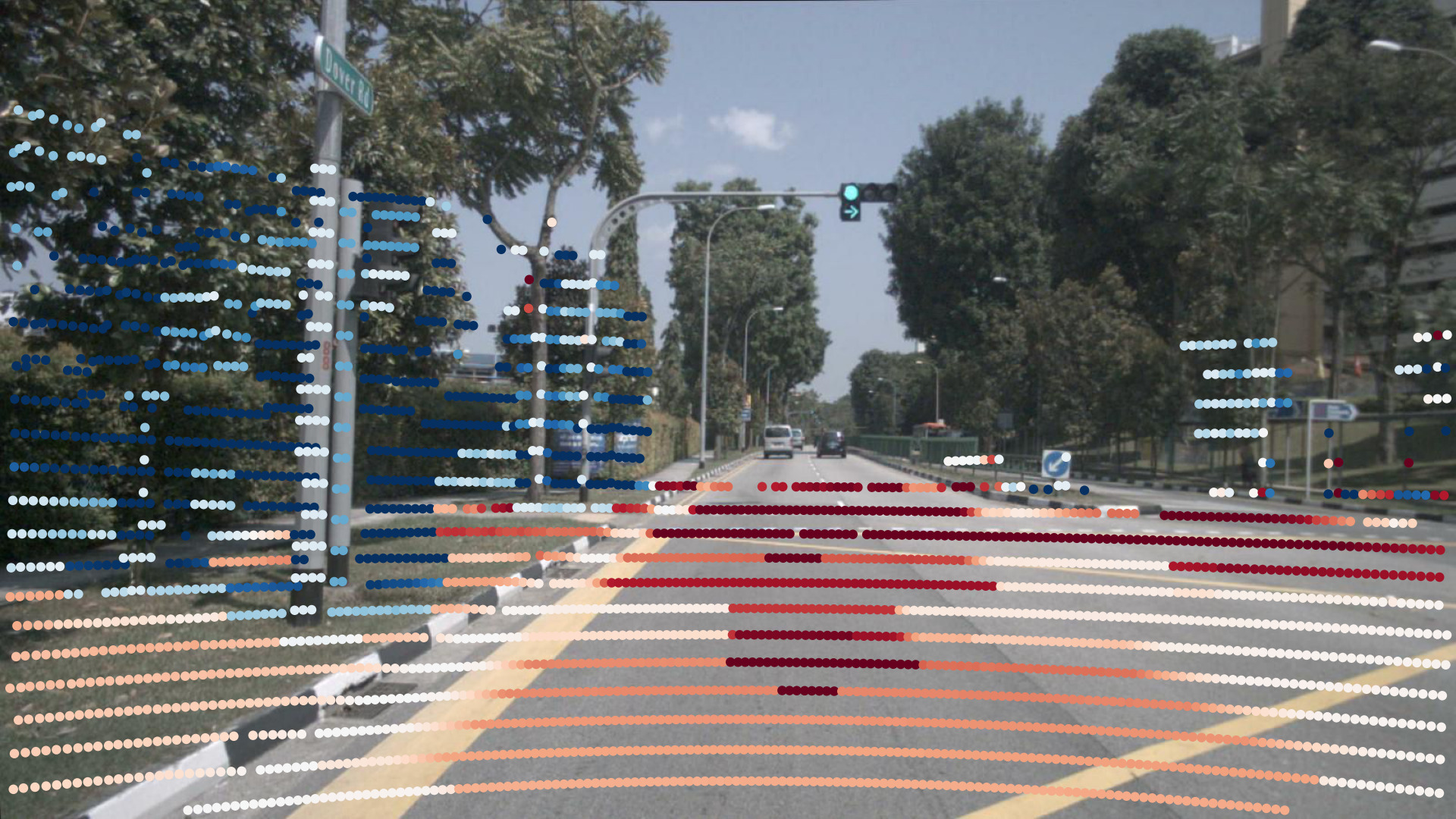};
				\nextgroupplot[title=Ours]
				\addplot graphics[xmax=1920, xmin=0, ymax=0, ymin=-1080]{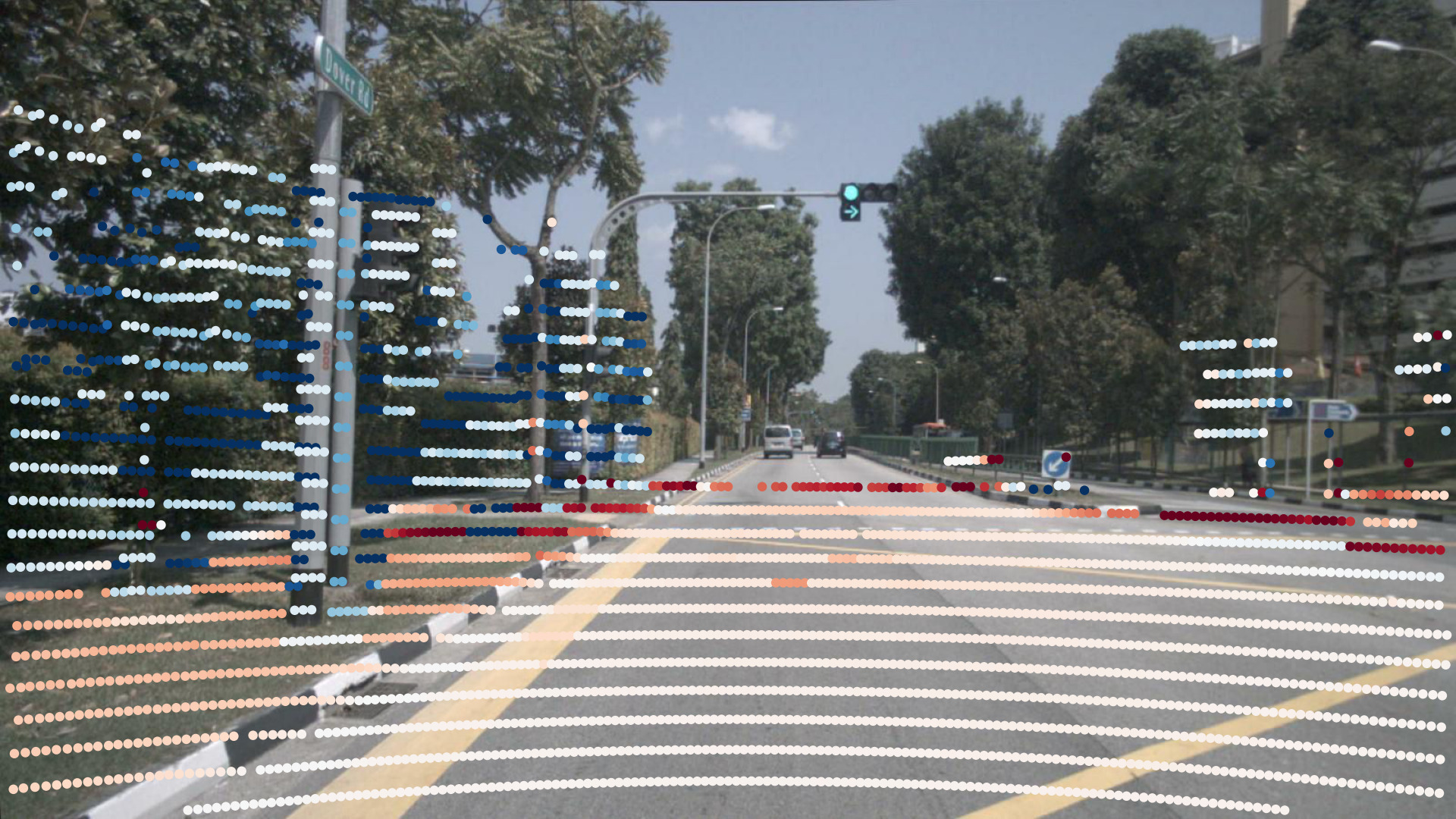};
				\nextgroupplot[
					colorbar horizontal,
					colorbar style={%
							anchor=north,
							at={(0.5,-0.07)},
							xlabel=Distance error / \unit{\metre},
						},
					title=Occ3D \cite{tian2023occ3d},
					ylabel=Model~\cite{huang2022bevdet4d},
				]

				\addplot graphics[xmax=1920, xmin=0, ymax=0, ymin=-1080]{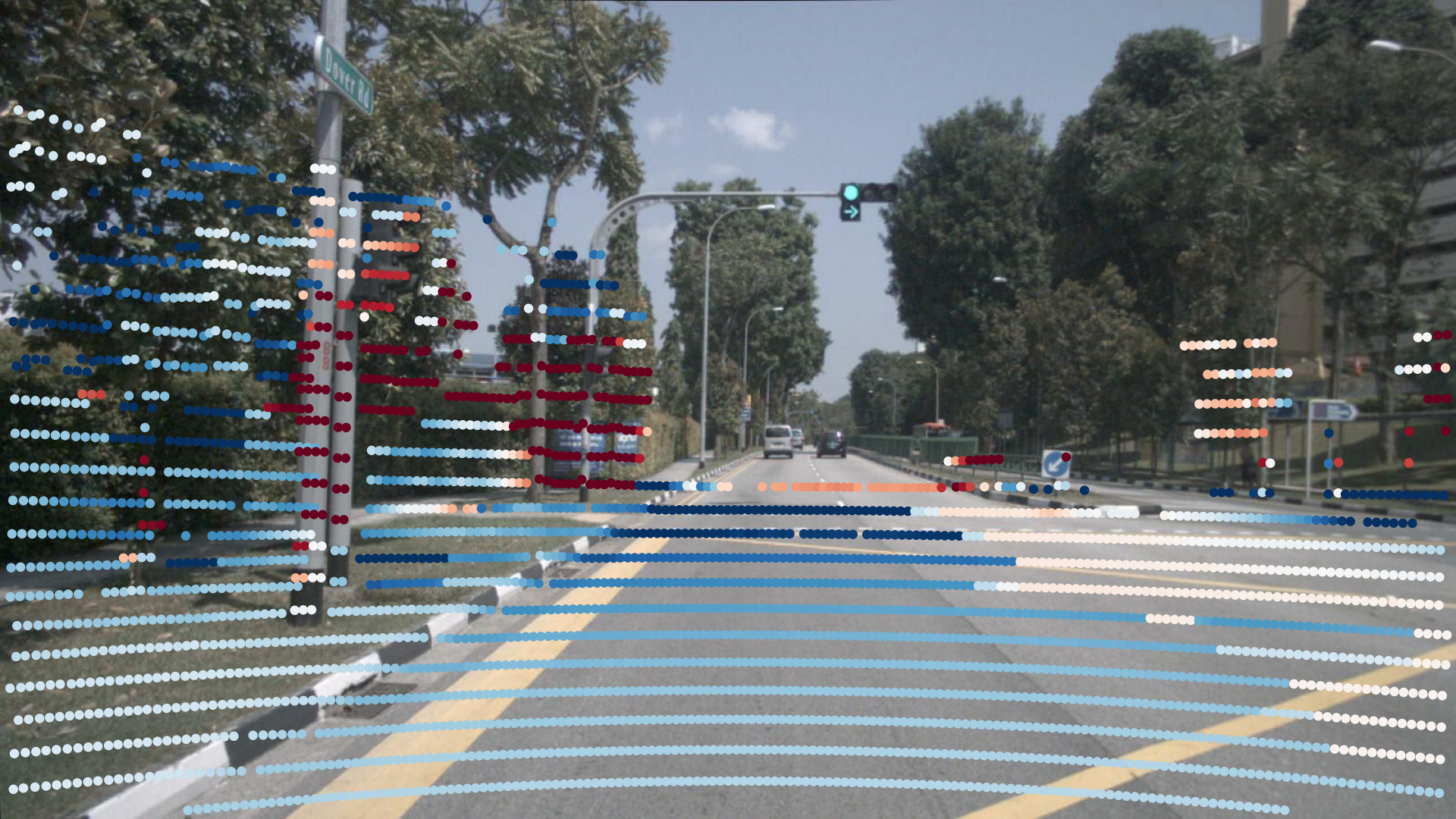};
				\nextgroupplot[
					colorbar horizontal,
					colorbar style={%
							anchor=north,
							at={(0.5,-0.07)},
							xlabel=Distance error / \unit{\metre},
						},
					title=Ours]
				\addplot graphics[xmax=1920, xmin=0, ymax=0, ymin=-1080]{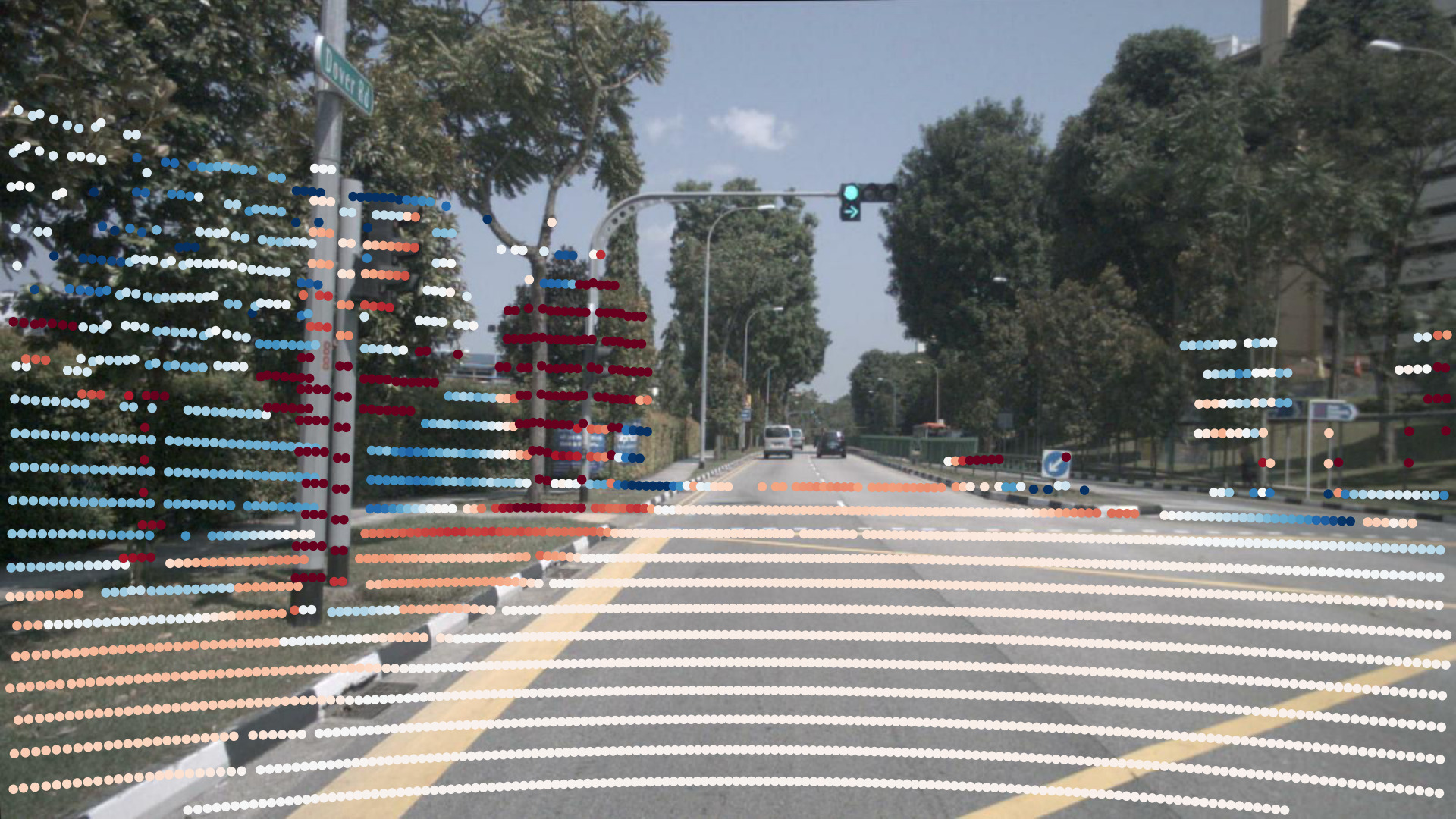};
			\end{groupplot}
		\end{tikzpicture}
	\end{adjustbox}
	\caption{%
		\textbf{Comparison of Depth Errors Between Occupancy Grid Maps and \acs{LIDAR} Measurements.}
		We compare our grid mapping approach (right) to Occ3D~\cite{tian2023occ3d} (left).
		Top: Depth errors between \acs{LIDAR} scan and ground-truth occupancy map.
		Bottom: Depth errors between \acs{LIDAR} scan and model predictions~\cite{huang2022bevdet4d}.
	}\label{fig:teaser}
\end{figure}

Enabling camera-based approaches for occupancy estimation requires data from real-world traffic scenes with \ac{GT} geometry, where \acs{LIDAR} is still the most reliable source of information due to its low measurement errors. First, refinement steps are applied to densify the sparse point clouds~\cite{tian2023occ3d, wang2023openocc}.
Then the \ac{GT} occupancy is reconstructed from the \acs{LIDAR} measurements and used to train and test occupancy prediction methods.
This is done by aggregating point clouds over multiple frames and compensating the movement of the ego vehicle and dynamic objects.
Voxels are considered occupied if they contain \ac{LIDAR} reflections and free if they are traversed by \ac{LIDAR} rays between sensor origin and surface. Errors in the reconstruction may occur due to multiple reasons: 1) noisy laser measurements, 2) few measurements available for a voxel, 3) incorrectly applied densification of points, and 4) incorrect motion compensation.

We show that the currently used techniques~\cite{tian2023occ3d, wang2023openocc,tong2023scene,wei2023surroundocc} yield very low-quality occupancy maps and present a new approach by modeling the sensor characteristics and using evidence theory to estimate the degree of belief in voxels being free, occupied, or uncertain.
Using this framework enables us to explicitly assign missing measurements, occlusions, or conflicts due to measurement, label or pose estimation errors to an \textit{``I don't know''} hypothesis.
The results show that our approach based on evidence theory is able to compensate for the reconstruction errors significantly better than existing methods, as it yields substantially more accurate occupancy maps when comparing them to the original \ac{LIDAR} measurements.
\cref{fig:teaser} shows an example of our method compared to Occ3D~\cite{tian2023occ3d}.

Furthermore, to train methods to predict occupancy from images, camera visibility masks are required in addition to the occupancy \acs{GT}.
We show that these masks significantly underestimate visibility and as a result prevent supervision in critical regions of the scene, manifesting in overestimated object boundaries and incorrect ground surface estimation.
Excluding all the regions that are incorrectly masked as invisible leads to methods being incapable of performing reliable estimates in these areas and eventually renders them unusable for safety-critical applications.
To this end, we propose a new loss weighting which makes use of our estimated uncertainty for supervision.
We show that using our data and loss weighting, the predictions from \acf{SotA} methods too can be substantially improved when comparing them to the original \ac{LIDAR} measurements.
Our contributions are threefold:
\begin{enumerate}
	\item We provide a simple, multi-frame 3D \ac{LIDAR} occupancy grid mapping method based on evidence theory, taking into account multiple input frames (see~\cref{sec:method}).
	      Our approach captures continuous evidence on voxels being occupied, free and uncertain due to measurement errors or missing observations from \acs{LIDAR} measurements.
	\item We show that the data provided by \acs{SotA} benchmarks~\cite{tian2023occ3d, tong2023scene} is inaccurate w.r.t.\ \acs{LIDAR} measurements and generate more accurate, evidential occupancy grid maps that can be used as a substitution for binary occupancy prediction of much higher quality.
	      Given our generated data, we demonstrate an occupancy estimation improvement of \( 30-53\% \) \ac{MAE} compared to the \acs{SotA} benchmark data (see~\cref{tab:rendered_lidar_depth}).
	\item Derived from our data, we propose a simple yet effective observability-based loss weighting to train \acs{SotA} models, producing \( 24\% \) lower \acp{MAE} between occupancy predictions and \acs{LIDAR} measurements (see~\cref{subsec:training}).
\end{enumerate}
The code to generate the data is available via \href{https://github.com/boschresearch/evidential-occupancy}{GitHub}\footnote{\href{https://github.com/boschresearch/evidential-occupancy}{https://github.com/boschresearch/evidential-occupancy}}.

\section{Related Work}\label{sec:related_work}

\paragraph{3D Occupancy Datasets.}
Datasets for automated driving, such as KITTI~\cite{geiger2012kitti}, nuScenes~\cite{caesar2019nuscenes}, Waymo~\cite{sun2019waymo} and Agroverse2~\cite{wilson2012argoverse2} record traffic scenes using multiple sensors like \acp{LIDAR} and cameras.
They provide additional data annotations to enable the training of  models for downstream tasks such as 3D object detection~\cite{zhiqi2022bevformer, li2023bevdepth, liu2023bevfusion, liu2022petr}.
3D occupancy prediction benchmarks for automated driving~\cite{tong2023scene, tian2023occ3d, li2023sscbench, wang2023openocc, wei2023surroundocc, wei2023surroundocc} use the recorded sensor data to generate dense 3D occupancy grids to represent arbitrary geometry of the scene surrounding the vehicle.

A basic processing pipeline to obtain dense occupancy grids consists of three steps:
First, \ac{LIDAR} measurements are aggregated over multiple time steps by considering static and dynamic objects separately.
Second, the points are partitioned into a dense voxel grid. Every voxel that contains a point is labeled as occupied.
Finally, occlusion reasoning is needed for non-occupied voxels to evaluate whether a voxel is free or unobserved.
Occ3D~\cite{tian2023occ3d} casts rays from the \ac{LIDAR} origins to each observed reflection and sets traversed voxels as free.
Additionally, for the tasks of vision-based occupancy prediction, \citeauthor{tian2023occ3d} calculate a camera visibility mask by casting rays from the camera origins to the centers of occupied voxels.
The first occupied voxel along each ray is marked as visible, any voxel behind it is labeled as unobserved due to occlusion.

Various techniques are used to improve the final voxel representation quality.
SurroundOcc~\cite{wei2023surroundocc} uses Poisson surface reconstruction~\cite{kazhdan2006poisson} to make the point clouds denser.
Occ3D~\cite{tian2023occ3d} proposes to fill the holes in the point cloud by mesh reconstruction through VDBFusion~\cite{vizzo2022vdbfusion} and use image-guided voxel refinement to enforce 2D-to-3D consistency between camera images and the occupancy grid.
OpenOccupancy~\cite{wang2023openocc} makes use of their Augment and Purifying pipeline together with additional human annotation effort to improve the label quality.
In contrast, our method proposes to embed information about free and unobserved space much earlier in the pipeline.
Therefore, we make use of the number of transmissions and reflections per voxel and resolve conflicting measurements using \acf{ET}.

\paragraph{3D Occupancy Prediction.}
The goal of occupancy prediction is to estimate the state of each voxel in a scene to be either \textit{occupied}, \textit{free} or \textit{unobserved} based on current and past camera images.
Semantic occupancy prediction additionally includes the goal of predicting a specific class for each occupied voxel.
The \textit{CVPR 2023 Occupancy Prediction Challenge} hosted by \citeauthor{tong2023scene} and \citeauthor{tian2023occ3d} focuses on the task of semantic occupancy prediction from camera images only.
The winning method FB-OCC by \citet{li2023fbbev, li2023fbocc} builds upon BEVDet~\cite{huang2021bevdet} and introduces forward-backward view transformations.
TPVFormer~\cite{huang2023tpvformer} uses three perpendicular planes to represent the 3D space efficiently.
BEVDepth~\cite{li2023bevdepth} introduces additional depth supervision to guide the process of lifting 2D image features into 3D space.
BEVDet4D~\cite{huang2022bevdet4d} exploits temporal information to improve 3D object detection.
Specifically, we choose BEVDet4D-Occ-R50 as our occupancy prediction baseline since it's widely adopted in the domain of vision based occupancy prediction and serves as baseline for many comparisons.

\paragraph{Occupancy Grid Mapping.}
Bayesian static occupancy grid mapping is a well-established concept initially developed by \citet{elfes1989using} and \citet{Moravec1989sensor} who model occupancy as a Bernoulli-distributed random variable.
These approaches, however, do not consider the amount of information collected to estimate the cell state, neither do they explicitly handle conflicts between different measurements.
\citet{yang2006evidential} resolve the above-mentioned issues using evidence theory.
Since then, more recent research focuses on 3D mapping, within dynamic environments and/or including semantics.
For instance, \citet{richter2023dual} model the dynamic traffic scene by two evidential semantic top-view grid maps, representing the ground and everything on top by two top-view grid maps.
\citet{baur2021slim} determine scene flow from \ac{LIDAR} measurements to improve mapping in dynamic environments.
In contrast to~\cite{baur2021slim, richter2023dual}, we focus on 3D evidential occupancy grid mapping, with instance flow determined by labeled object shapes and motion.

\section{Method}\label{sec:method}

Our method starts by mapping \ac{LIDAR} measurements into a spherical voxel grid by adding up the reflections and transmissions for each voxel.
In the next step, we warp the spherical voxel grids from different time steps to a common Cartesian voxel grid for the current time step and aggregate the number of reflections and transmissions.
Finally, we assign evidences to a voxel being \textit{occupied}, \textit{free} or \textit{uncertain}.
In the following, we describe each step in detail.

\begin{figure*}[t]
    \centering
    \def\svgwidth{\textwidth}
    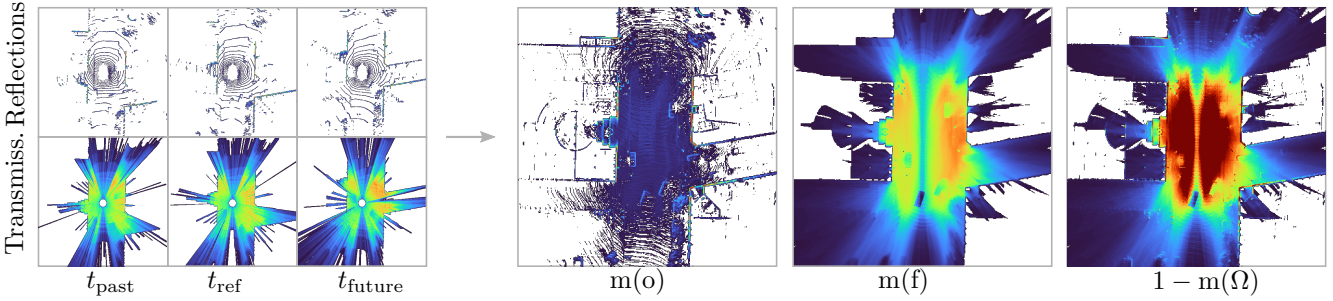
    \caption{%
        \textbf{Conversion of Transmissions and Reflections to Beliefs.}
        We start by calculating reflections and transmissions in a spherical coordinate system for each time step individually (left). 
        Then we aggregate the number of reflections and transmissions from past, reference, and future frames into one common Cartesian grid at the reference time $t_\text{ref}$.
        During this process, we compensate for object and ego vehicle motion.
        We use the basic belief assignment described in \cref{sec:bba} to obtain the occupied belief $\func{m}`*(\text{o})$, the free belief $\func{m}`*(\text{f})$, and the observability $1 - \func{m}`*(\Omega)$.
    }\label{fig:method_overview}
\end{figure*}

\subsection{Spherical Sensor Mapping}\label{subsec:spherical_mapping}
We start by mapping the \ac{LIDAR} measurements from every frame into two spherical grid maps denoted as \textit{reflections} and \textit{transmissions}.
Every voxel in the reflections grid map encodes how many reflections originated from there, which is usually high for voxels capturing object surfaces.
Every voxel in the transmissions grid map, on the other hand, encodes how many \ac{LIDAR} rays freely passed this voxel from the \ac{LIDAR} origin to their reflection positions, which is usually high for voxels representing free space.

\begin{figure}
    \begin{tikzpicture}
        \begin{groupplot}[
                enlargelimits=0,
                grid,
                grid style={%
                        thick,
                        black,
                    },
                group style={%
                        group size=1 by 2,
                        vertical sep=5mm,
                    },
                height={},
                samples=4,
                shader=flat,
                tickwidth=0mm,
                title style={%
                        at={(0.99, 0.87)},
                    },
                view={0}{90},
                x label style={at={(axis description cs:0.5,0)}},
                xlabel={Distance \( \rho \rightarrow \)},
                xmax=7.5001,
                xmin=0,
                xtick={0, 1, ..., 8},
                xticklabels={, ,},
                y label style={at={(axis description cs:0,0.5)}},
                ylabel={Azimuth \( \phi \rightarrow \)},
                ymax=3.0001,
                ymin=0,
                ytick={0, 1, ..., 3},
                yticklabels={, ,},
                zmin=0,
                zmax=2.3,
            ]

            \nextgroupplot[
                colorbar,
                colorbar style={%
                        ymin=0,
                        ytick={0, 0.25, ..., 1},
                    },
                height=35mm,
                title=Reflections,
            ]
            \addplot[black, mark=*, thin, red] coordinates {(1.5,1.5)};
            \addplot[black, mark=*, thin, red] coordinates {(3.9,1.7)};
            \addplot[black, mark=*, thin, red] coordinates {(6,1)};
            \addplot3[surf,domain=1:2,y domain=1:2] {1};
            \addplot3[surf,domain=3:4,y domain=1:2] {0.48};
            \addplot3[surf,domain=3:4,y domain=2:3] {0.12};
            \addplot3[surf,domain=4:5,y domain=1:2] {0.32};
            \addplot3[surf,domain=4:5,y domain=2:3] {0.08};
            \addplot3[surf,domain=5:6,y domain=0:1] {0.25};
            \addplot3[surf,domain=5:6,y domain=1:2] {0.25};
            \addplot3[surf,domain=6:7,y domain=0:1] {0.25};
            \addplot3[surf,domain=6:7,y domain=1:2] {0.25};
            \draw[densely dotted] (1,1) rectangle (2,2);
            \draw[densely dotted] (3.4,1.2) rectangle (4.4,2.2);
            \draw[densely dotted] (5.5,0.5) rectangle (6.5,1.5);
            \draw[densely dotted, red, very thick] (0,1.5) -- (1.5,1.5);
            \draw[densely dotted, red, very thick] (0,1.7) -- (3.9,1.7);
            \draw[densely dotted, red, very thick] (0,1) -- (6,1);
            \node[font=\tiny, text opacity=0.6] at (1.7,1.25) {1};
            \node[font=\tiny, text opacity=0.6] at (3.7,1.25) {0.48};
            \node[font=\tiny, text opacity=0.6] at (3.7,2.25) {0.12};
            \node[font=\tiny, text opacity=0.6] at (4.7,1.25) {0.32};
            \node[font=\tiny, text opacity=0.6] at (4.7,2.25) {0.08};
            \node[font=\tiny, text opacity=0.6] at (5.7,0.25) {0.25};
            \node[font=\tiny, text opacity=0.6] at (6.7,0.25) {0.25};
            \node[font=\tiny, text opacity=0.6] at (6.7,1.25) {0.25};
            \node[font=\tiny, text opacity=0.6] at (5.7,1.25) {0.25};

            \nextgroupplot[
                colorbar,
                colorbar style={%
                        ymin=0,
                    },
                height=35mm,
                title=Transmissions,
            ]
            \addplot[black, mark=*, red, thin] coordinates {(1.5,1.5)};
            \addplot[black, mark=*, red, thin] coordinates {(3.9,1.7)};
            \addplot[black, mark=*, red, thin] coordinates {(6,1)};
            \addplot3[surf,domain=3:4,y domain=2:3] {0.08};
            \addplot3[surf,domain=0:3,y domain=2:3] {0.12 + 0.08};
            \addplot3[surf,domain=5:6,y domain=1:2] {0.25};
            \addplot3[surf,domain=4:5,y domain=1:2] {0.25 + 0.25};
            \addplot3[surf,domain=3:4,y domain=1:2] {0.32 + 0.25 + 0.25};
            \addplot3[surf,domain=1:3,y domain=1:2] {0.48 + 0.32 + 0.25 + 0.25};
            \addplot3[surf,domain=0:1,y domain=1:2] {1 + 0.48 + 0.32 + 0.25 + 0.25};
            \addplot3[surf,domain=5:6,y domain=0:1] {0.25};
            \addplot3[surf,domain=0:5,y domain=0:1] {0.25 + 0.25};
            \draw[densely dotted, very thick, red] (0,1.5) -- (1.5,1.5);
            \draw[densely dotted, very thick, red] (0,1.7) -- (3.9,1.7);
            \draw[densely dotted, very thick, red] (0,1) -- (6,1);
            \node[font=\tiny, text opacity=0.6] at (0.7,0.25) {0.5};
            \node[font=\tiny, text opacity=0.6] at (0.7,1.25) {2.3};
            \node[font=\tiny, text opacity=0.6] at (0.7,2.25) {0.2};
            \node[font=\tiny, text opacity=0.6] at (1.7,0.25) {0.5};
            \node[font=\tiny, text opacity=0.6] at (1.7,1.25) {1.3};
            \node[font=\tiny, text opacity=0.6] at (1.7,2.25) {0.2};
            \node[font=\tiny, text opacity=0.6] at (2.7,0.25) {0.5};
            \node[font=\tiny, text opacity=0.6] at (2.7,1.25) {1.3};
            \node[font=\tiny, text opacity=0.6] at (2.7,2.25) {0.2};
            \node[font=\tiny, text opacity=0.6] at (3.7,0.25) {0.5};
            \node[font=\tiny, text opacity=0.6] at (3.7,1.25) {0.82};
            \node[font=\tiny, text opacity=0.6] at (3.7,2.25) {0.08};
            \node[font=\tiny, text opacity=0.6] at (4.7,0.25) {0.5};
            \node[font=\tiny, text opacity=0.6] at (4.7,1.25) {0.5};
            \node[font=\tiny, text opacity=0.6] at (5.7,0.25) {0.25};
            \node[font=\tiny, text opacity=0.6] at (5.7,1.25) {0.25};
        \end{groupplot}
    \end{tikzpicture}
    \caption{%
        \textbf{Spherical Reflection and Transmission Grid Mapping.}
        Top: Weighted scattering of measurements to spherical voxels.
        Bottom: Transmission value computation through cumulative sum starting from maximum and ending in minimum distance.
        For illustration, we omit the polar angle \( \theta \).
        Colors denote the value assigned to a grid cell.
        The bottom shows the corresponding transmissions that are derived from the reflections.
    }\label{fig:grid_mapping}
\end{figure}
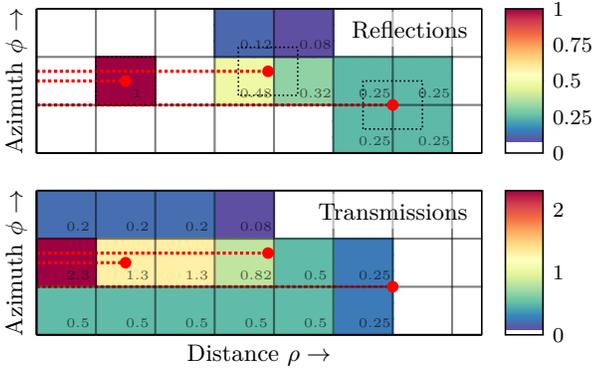

Given the \ac{LIDAR} \ac{FOV} and its characteristics, we map measured points into spherical coordinates with radial distance (\( \rho \)), polar angle (\( \theta \)), and azimuth angle (\( \phi \)), which we then quantize with \( \Delta \rho, \Delta \theta, \Delta \phi \).
The spherical voxel size corresponds to the measurement errors \( \pm \frac{\Delta \rho}{2} \), \( \pm \frac{\Delta \theta}{2} \) and \( \pm \frac{\Delta \phi}{2} \) that we obtain from the sensor characteristics.

\paragraph{Reflections Grid Map.}
As illustrated in the top of~\cref{fig:grid_mapping}, to obtain the reflection value for each voxel, we define a cube centered at the current measurement point with the same size as the grid cells.
We then use the volume overlap to distribute the probability mass of the measurement into the neighboring voxels.
This step can be efficiently implemented as weighted point scattering.
As a result, each spherical voxel (\( \rho, \phi, \theta \)) is assigned a reflection value \( r \).

\paragraph{Transmissions Grid Map.}
Given the reflections, we can now compute the transmissions grid map.
For a given (\( \rho, \phi, \theta \)) bin, its transmission
\begin{equation}
	q(\rho, \phi, \theta) = \sum_{\rho' > \rho} r(\rho', \phi, \theta)
\end{equation}
is determined as the sum of all reflection values that lie within the same \( \phi \) and \( \theta \) bin but have a distance larger than \( \rho \).
Intuitively, this means that the more rays have landed farther than the current bin and therefore must have passed though the current bin, the higher the likelihood that this bin is free.
\cref{fig:grid_mapping} illustrates an example of this computation step which can be efficiently implemented via the exclusive cumulative sum of reflections from the maximum to the minimum radial distance.

Note that generally \( r \) and \( q \) are not probability values, but the number of events associated to a voxel, taking any positive real value due to the weighted assignment.

\subsection{Multi-Frame Aggregation}

In this step, we aggregate the reflections and transmissions from multiple time steps into a single Cartesian voxel grid and warp information from past and future frames into the reference frame using the known per-object motion.
Here, we consider all frames within a viewpoint displacement due to ego motion of less than \( \delta_\text{max} \) from the reference frame at time \( t_\text{ref} \), limited to at most \( N_\text{max} \) frames evenly sampled in time.

We start by defining a Cartesian volume aligned to the current vehicle pose in the reference frame at time \( t_\text{ref} \).
We consider the coordinate system of this frame to be the target coordinate system, into which all other frames (at different \( t \)) are warped.
We assume that for every annotated object \( o \) from the reference frame we know its scene flow \( f_o(t) \) with every past/future frame \( t \).
\(f_o(t)\) allows us to transform every object point \(p\) from the reference frame \(t_\text{ref}\) into frame \(t\) as follows:
\begin{equation}
	p(\tilde{t}) = f_o(\tilde{t}) + p(t) \,.
\end{equation}

We now iterate over every object in frame \(t_\text{ref}\) and transform every of its \(i\)-th voxel centers \(p_i(t_\text{ref})\) into all future and past frames, yielding a set of \(T\) new coordinates \( \{p_i(t)\}_{t=1}^{T}\).
Note that here \(t_\text{ref}\) is included among the \(T\) time steps, with all flow vectors being zero.
Next, we transform each coordinate \(p_i(t) = [x_i(t), y_i(t), z_i(t)]\) into the spherical coordinates
\begin{align}
	\rho_i(t)   & = \sqrt{x_i(t)^2 + y_i(t)^2 + z_i(t)^2} \quad \text{with} \\
	\theta_i(t) & = \arccos`*(\frac{z_i(t)}{\rho_i(t)})  \, \text{and}   \,
	\phi_i(t)    = \arctan`*(\frac{y_i(t)}{x_i(t)}) \, .\notag
\end{align}
Finally, we retrieve the number of reflections and transmissions via trilinear interpolation from the corresponding precomputed maps.
In addition, we normalize sampled values by multiplying with the voxel volume ratio
\begin{align}
	s                  & = \frac{V_\text{cartesian}}{V_\text{spherical}} \quad\mathrm{with}                                                                                                                                                                                                                                                                                         \\
	V_\text{cartesian} & = \Delta x \Delta y \Delta z   \quad\mathrm{and}                                                                                                                                                                                                                                                                                                           \\
	V_\text{spherical} & = \int \limits_{\rho - \frac{\Delta \rho}{2}}^{\rho + \frac{\Delta \rho}{2}} \int \limits_{\theta - \frac{\Delta \theta}{2}}^{\theta + \frac{\Delta \theta}{2}} \int \limits_{\phi - \frac{\phi r}{2}}^{\phi + \frac{\Delta \phi}{2}} \tilde{\rho}^2 \sin `*(\tilde{\theta})\ \mathrm{d}\tilde{\rho}\ \mathrm{d}\tilde{\theta}\ \mathrm{d}\tilde{\phi} \,,
\end{align}
between Cartesian and spherical cells.
Finally, we aggregate the number of reflections and transmissions
\begin{equation}
	r_i = \frac{1}{T} \sum_{t=1}^{T} r_i(t) \quad\mathrm{and}\quad q_i = \frac{1}{T} \sum_{t=1}^{T} q_i(t)
\end{equation}
from all \( T \leq N_\text{max} \) frames within \( \delta_\text{max} \), normalized by the number of considered frames.

\subsection{Evidential Mapping}\label{sec:bba}

Each voxel in the final grid map should not only represent information about occupancy and free space, but also uncertainty.
To determine this information, we use the framework of \ac{ET} (also referred to as \ac{DST}), which can be interpreted as a generalization of Bayesian theory~\cite{dempster1968theory}.
It allows assigning evidence from different information sources not only to a single hypothesis but to all possible combinations.
Let \( \Omega \) be the set of hypotheses under consideration.
Then, the \ac{BBA}
\begin{equation}
	\func{m} \colon 2^{\Omega} \rightarrow `*[0, 1]\,, \quad \bba{\emptyset} = 0\,, \quad \sum_{X \in 2^\Omega} \bba{X} = 1 \label{eq:bba}
\end{equation}
assigns evidence mass to each element of its power set \( 2^\Omega \).
Here, we define
\begin{equation}
	\Omega = `{\text{o}, \text{f}} \,,
\end{equation}
containing the hypotheses \textit{occupied} (o) and \textit{free} (f).
Note that the hypothesis \( \Omega \) contains both o and f and corresponds to being uncertain.
As suggested in~\cite{richter2023dual}, we use the \ac{BBA}
\begin{equation}
	\func{m}`*(\omega | q, r) =
	\begin{dcases*}
		p_\text{FN}^q `*(1 - p_\text{FP}^r)             & for \( \omega=\text{o} \) \\
		p_\text{FP}^r `*(1 - p_\text{FN}^q)             & for \( \omega=\text{f} \) \\
		1 - \func{m}`*(\text{o}) - \func{m}`*(\text{f}) & for \( \omega=\Omega \)
	\end{dcases*}\label{eq:occupancy_bba}
\end{equation}
to determine each voxel's evidence towards every possible hypothesis \( \omega \).

In this formulation, the false negative probability \( p_\text{FN} \) and false positive probability \( p_\text{FP} \) are parameters characterizing the sensor.
\( p_\text{FN} \) is the probability of a voxel being estimated as free when it should in fact be occupied, \( p_\text{FP} \) is the probability of predicting the state of a voxel to be occupied when it should be free.
These values are fixed for the entire dataset, and are determined heuristically as described in section~\cref{sec:hyperparam}. We show an example of input reflections and transmissions and our estimated beliefs in~\cref{fig:method_overview}.

\section{Experiments}\label{sec:experiments}

\subsection{Evaluation}\label{sec:evaluation}

We have observed that occupancy grids and especially visibility masks, which are used for model training, are of low quality.
Therefore, we propose to quantify the agreement between the raw \ac{LIDAR} measurements and the reconstructed occupancies by rendering depth maps and comparing them.
We argue that comparing to the raw \ac{LIDAR} measurements provides the most reliable available reference.

\paragraph{\Ac{BBA} to Binary Occupancy.}
To compare our continuous occupancy maps to other occupancy prediction methods, we need a binary occupancy value per voxel.
Therefore, we compare the occupied belief \(\func{m_i}`*(\text{o})\) and the free belief \(\func{m_i}`*(\text{f})\) and assign a binary occupancy value \(o_i\) to each voxel \(i\)
\begin{equation}
	o_i =
	\begin{dcases*}
		1 & if \(\func{m_i}`*(\text{o}) > \func{m_i}`*(\text{f})\) \\
		0 & else
	\end{dcases*}.
	\label{eq:bba_binary}
\end{equation}
This yields a binary occupancy map, which we use to compare against other occupancy methods.

\paragraph{Depth Evaluation.}
For each existing \acs{GT} \ac{LIDAR} point \(i\), we march along the corresponding ray and extract the predicted depth value \(\text{d}^{pr}_i\) as the first intersection point of this ray with a voxel surface predicted to be occupied.
We then compare the predicted value \(\text{d}^{pr}_i\) with the \acs{GT} \ac{LIDAR} depth \(\text{d}^{gt}_i\).
We report several standard metrics from the depth estimation literature: \ac{MAE}, \ac{RMSE} and its log-variant (\ac{RMSE}\(_{log}\)), and the percentage of points that have a relative depth deviation \(\delta\) smaller than a specified threshold.
Since the dense occupancy grid has a limited lower and upper bound, we only consider \ac{LIDAR} measurements within the enclosed volume. A complete overview of used parameters is given in~\cref{tab:params}.

\subsection{Occupancy Grid Comparison}
We compare our generated occupancy grids to those of Scene as Occupancy~\cite{tong2023scene}, SurroundOcc~\cite{wei2023surroundocc}, Occ3D~\cite{tian2023occ3d} and OpenOccupancy~\cite{wang2023openocc} on nuScenes~\cite{caesar2019nuscenes} and to those of Occ3D~\cite{tian2023occ3d} on  Waymo~\cite{sun2019waymo}.
Especially, Occ3D provides the semantic occupancy labels for the \textit{CVPR 2023 3D Occupancy Prediction Challenge} built on top of the nuScenes dataset~\cite{caesar2019nuscenes}.

\begin{figure*}[t]
	\centering
	\includegraphics[width=\linewidth]{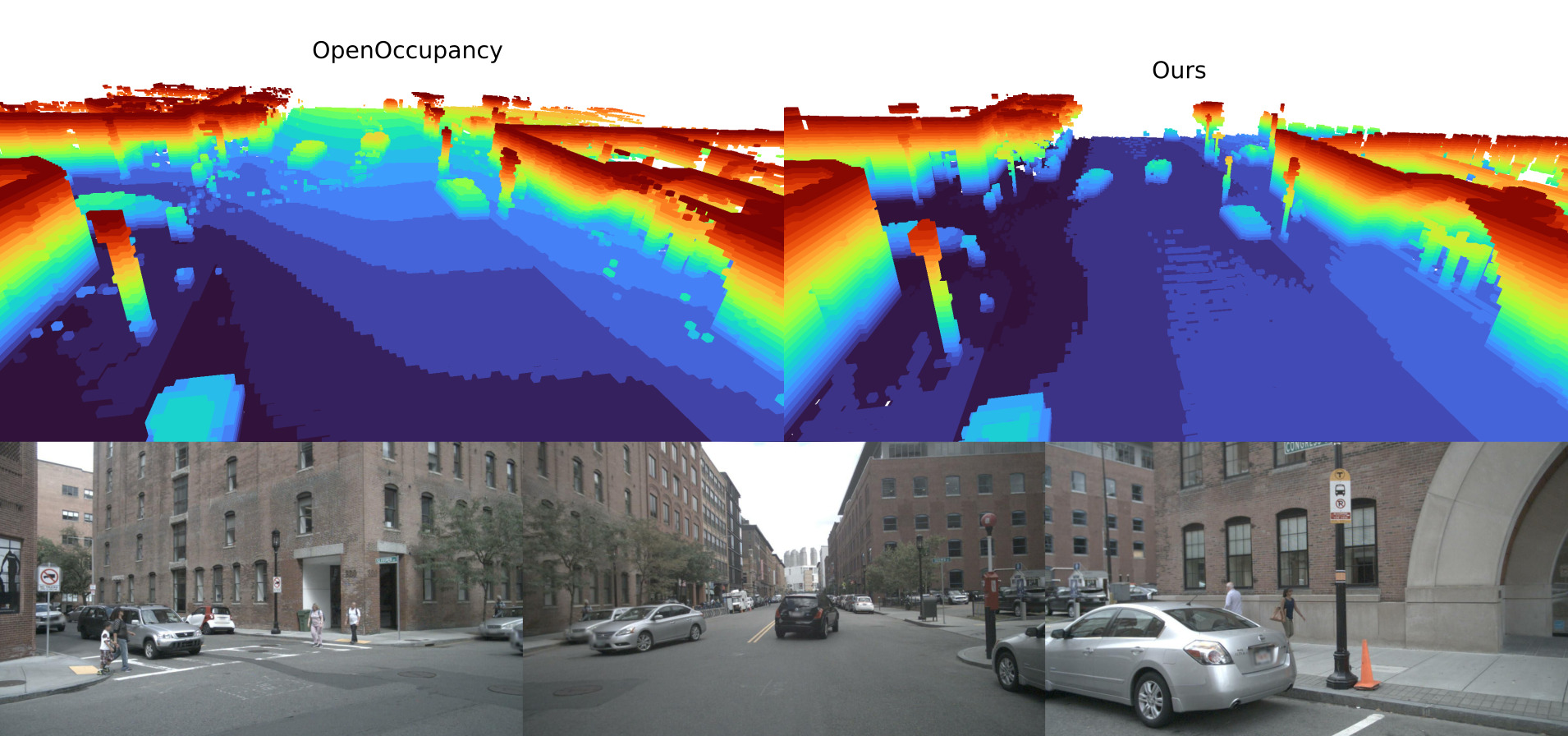}
	\caption{
		\textbf{Comparison of Occupancy Maps.}
		We show the occupancy map by OpenOccupancy~\cite{wang2023openocc} (left) and ours (right) for a nuScenes~\cite{caesar2019nuscenes} scene. Both methods use a voxel size of \qty{0.2}{\metre}. We encode the color depending on the voxel's \(z\)-coordinate. The bottom row depicts the three front facing cameras of the vehicle. Our method shows improved quality on the ground surface and thin objects like the streetlights. It also reduces flying particles behind moving vehicles and pedestrians.}
	\label{fig:occupancy_gt}
\end{figure*}

We use the described evaluation protocol from \cref{sec:evaluation} and compare rendered depths in the occupancy grids to the \ac{LIDAR} measurements.
We summarize the quantitative results in~\cref{tab:rendered_lidar_depth} and show a qualitative result of the rendered depth error in~\cref{fig:teaser}.
\cref{fig:occupancy_gt} depicts qualitatively the improvement of our method.
The ground surface is significantly more accurate and objects like the streetlights have sharper boundaries.
Flying artifacts behind moving objects are reduced.
The supplementary material contains additional qualitative results.
Our method outperforms all baseline methods across all metrics by a large margin.
Therefore, our generated occupancy maps matches the \ac{LIDAR} measurements better than previous methods according to all evaluated metrics.
This also holds for different voxel sizes and across different datasets:
We outperform other methods~\cite{tong2023scene, wei2023surroundocc, tian2023occ3d, wang2023openocc} with comparable voxel size on nuScenes~\cite{caesar2019nuscenes} and Waymo~\cite{sun2019waymo}.

\begin{table*}[t]
	\centering
	\begin{tabular}{l l r r r r r r r}
		\toprule
		                                                    & Method @ Voxel size                                         & \acs{MAE} \lowerbetter & \acs{RMSE} \lowerbetter & \acs{RMSE}$_{log}$ \lowerbetter & \( \delta < 1.25 \) \higherbetter & \( \delta < 1.25^2 \) \higherbetter & \( \delta < 1.25^3 \) \higherbetter \\
		                                                    &                                                             & in \unit{\metre}       & in \unit{\metre}        & in \( \log{\unit{\metre}} \)    & in \%                             & in \%                               & in \%                               \\
		\midrule
		\multirow{6}{*}{\rotatebox[origin=c]{90}{nuScenes}} & Scene as Occupancy~\cite{tong2023scene} @ \qty{0.5}{\metre} & 2.44                   & 4.99                    & 0.57                            & 64.4                              & 81.7                                & 87.4                                \\
		                                                    & SurroundOcc~\cite{wei2023surroundocc}   @ \qty{0.5}{\metre} & 2.36                   & 5.11                    & 0.46                            & 72.0                              & 83.6                                & 89.0                                \\
		                                                    & Occ3D~\cite{tian2023occ3d}              @ \qty{0.4}{\metre} & 1.80                   & 4.43                    & 0.37                            & 81.4                              & 89.7                                & 93.3                                \\
		                                                    & Ours                                    @ \qty{0.4}{\metre} & \ranktwo{1.25}         & \ranktwo{3.42}          & \ranktwo{0.28}                  & \ranktwo{87.9}                    & \ranktwo{94.6}                      & \ranktwo{96.8}                      \\
		                                                    & OpenOccupancy~\cite{wang2023openocc}    @ \qty{0.2}{\metre} & 1.89                   & 4.39                    & 0.51                            & 75.0                              & 85.5                                & 90.5                                \\
		                                                    & Ours                                    @ \qty{0.2}{\metre} & \rankone{0.91}         & \rankone{2.99}          & \rankone{0.24}                  & \rankone{92.6}                    & \rankone{96.2}                      & \rankone{97.7}                      \\
		\midrule
		\multirow{3}{*}{\rotatebox[origin=c]{90}{Waymo}}    & Occ3D~\cite{tian2023occ3d}              @ \qty{0.4}{\metre} & 3.23                   & 6.52                    & 0.56                            & 75.6                              & 83.1                                & 87.6                                \\
		                                                    & Ours                                    @ \qty{0.4}{\metre} & \ranktwo{1.51}         & \ranktwo{3.98}          & \ranktwo{0.23}                  & \ranktwo{90.6}                    & \ranktwo{95.1}                      & \ranktwo{97.1}                      \\
		                                                    & Ours                                    @ \qty{0.2}{\metre} & \rankone{1.15}         & \rankone{3.96}          & \rankone{0.21}                  & \rankone{92.8}                    & \rankone{95.6}                      & \rankone{97.3}                      \\
		\bottomrule
	\end{tabular}
	\caption{%
		\textbf{Comparison of Depth Errors Between Occupancy Grid Maps and \acs{LIDAR} Measurements.}
		We use raw \ac{LIDAR} measurements as a common ground truth to compare between different occupancy grid maps.
		Therefore, we calculate the first intersection of a \ac{LIDAR} ray and an occupied voxel to determine a depth value as described in \cref{sec:evaluation}.
		We use established depth estimation metrics to evaluate the quality of the generated occupancy maps.
		For~\cite{tong2023scene,wei2023surroundocc,tian2023occ3d,wang2023openocc}, we map all semantic classes to be occupied.
		We use our \acs{BBA} to determine occupied and free voxel cells by comparing the occupied belief \(\func{m}`*(\text{o})\) and free belief \(\func{m}`*(\text{f})\).
		Apart from the different spherical grid extents due to different \acsp{FOV} of the \acs{LIDAR} sensors we keep all hyperparameters constant for both datasets.
		We outperform existing methods with comparable voxel size across all metrics by a large margin both on nuScenes~\cite{caesar2019nuscenes} and on Waymo~\cite{sun2019waymo}.
	}%
	\label{tab:rendered_lidar_depth}
\end{table*}

\subsection{Hyperparameters}\label{sec:hyperparam}

\begin{table}
	\centering
	\small
	\begin{adjustbox}
		{max width=\columnwidth}
		\begin{tabular}{l l}
			\toprule
			Parameters              & Value                                                                                  \\
			\midrule
			Spherical grid extent   & \( \rho \) = \SIrange{2.5}{60}{\metre}                                                 \\
			                        & \( \theta \) = \SIrange{75}{125}{\degree}                                              \\
			                        & \( \phi \) = \SIrange{-180}{180}{\degree}                                              \\
			Spherical cell size     & \( \Delta \rho = \qty{0.1}{\metre},\ \Delta \theta = \Delta \phi = \ang{0.5} \)        \\
			Cartesian grid extent   & \( x \) = \SIrange{-40}{40}{\metre}                                                    \\
			                        & \( y \) = \SIrange{-40}{40}{\metre}                                                    \\
			                        & \( z \) = \SIrange{-1}{5.4}{\metre}                                                    \\
			Cartesian cell size     & \( \Delta x = \Delta y = \Delta z = \qty{0.2}{\metre} \text{ or } \qty{0.4}{\metre} \) \\
			False neg.\ probability & \( p_\text{FN} = 0.8 \text{ or } 0.9 \)                                                \\
			False pos.\ probability & \( p_\text{FP} = 0.2 \text{ or } 0.1 \)                                                \\
			Max.\ number of frames   & \(N_\text{max} = 50\)                                                                               \\
			Max.\ displacement      & \(\delta_\text{max} = \qty{20}{\metre}\)                                                                   \\
			\bottomrule
		\end{tabular}
	\end{adjustbox}
	\caption{\textbf{Parameter Overview.} If not noted otherwise, we use the parameters listed here. We choose the spherical grid extent such that it encloses most of the \ac{LIDAR} measurements. The Cartesian grid extent matches the one of~\cite{tian2023occ3d}. }
	\label{tab:params}
\end{table}

\begin{figure}[htbp]
	\begin{tikzpicture}
		\pgfplotstableread{data/heatmap/mini_p_fn_p_fp_sweep.csv}\dataTwenty;
		\pgfplotstableread{data/heatmap/mini_p_fn_p_fp_sweep_04.csv}\dataForty;
		\begin{groupplot}[
				axis equal image,
				group style={%
						group size=2 by 2,
						horizontal sep=3mm,
						vertical sep=3mm,
					},
				point meta=explicit,
				title style={%
						anchor=north,
						at={(0.5, 0.95)},
					},
				width=0.7\columnwidth,
				xlabel=\(p_\text{FN}\),
				xtick={0, 0.2, ..., 1},
				ylabel=\(p_\text{FP}\),
				ytick={0, 0.2, ..., 1},
			]
			\nextgroupplot[title=\Acs{MAE} @ \qty{20}{\cm}]
			\addplot [matrix plot*] table
				[
					x=p_fn,
					y=p_fp,
					meta=mae,
				]
				{\dataTwenty};
			\nextgroupplot[title=\(\delta \leq 1.25\) @ \qty{20}{\cm}]
			\addplot [matrix plot*] table
				[
					x=p_fn,
					y=p_fp,
					meta=delta_1_25,
				]
				{\dataTwenty};
			\nextgroupplot[title=\Acs{MAE} @ \qty{40}{\cm}]
			\addplot [matrix plot*] table
				[
					x=p_fn,
					y=p_fp,
					meta=mae,
				]
				{\dataForty};
			\nextgroupplot[title=\(\delta \leq 1.25\) @ \qty{40}{\cm}]
			\addplot [matrix plot*] table
				[
					x=p_fn,
					y=p_fp,
					meta=delta_1_25,
				]
				{\dataForty};
		\end{groupplot}
	\end{tikzpicture}
	\caption{
 \textbf{Determining the False Negative and False Positive Probabilities.}
		To select reasonable values for \( p_\text{FP} \) and \( p_\text{FN} \), we run a grid search on the nuScenes mini split and evaluate against the raw \ac{LIDAR} data.
        We plot \ac{MAE}~(\lowerbetter , left) and relative depth deviation accuracy \( \delta < 1.25 \)~(\higherbetter , right) for a voxel size of \qty{20}{\cm} and \qty{40}{\cm} (top and bottom row).
        Based on the data, we set \( p_\text{FP} = 0.2 \,\,\mathrm{ and }\,\, p_\text{FN} = 0.8 \) for a voxel size of \qty{20}{\cm}, and \( p_\text{FP} = 0.1 \,\,\mathrm{ and }\,\, p_\text{FN} = 0.9 \) for a voxel size of \qty{40}{\cm}.
	}\label{fig:heatmap}
\end{figure}
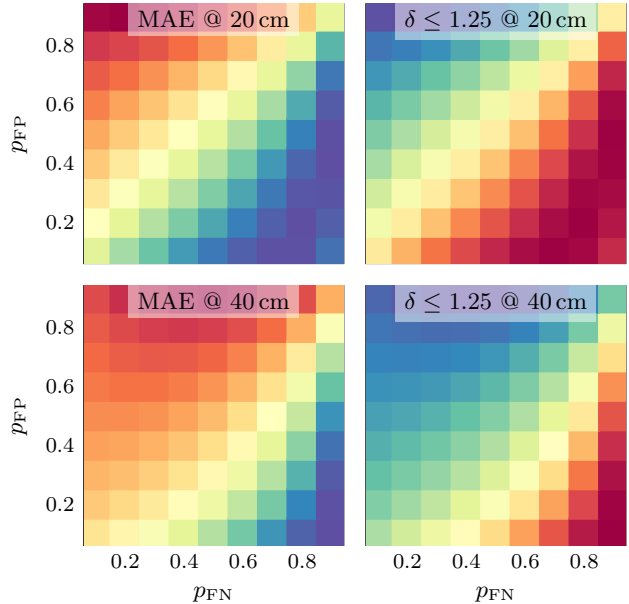

As a first step, we determine the values for the false negative and false positive probabilities \( p_\text{FN} \) and \( p_\text{FP} \) from~\cref{eq:occupancy_bba}.
These probabilities depend on multiple factors, including sensor characteristics and hyperparameters like the Cartesian voxel cell size.
Therefore, we do a grid search over possible combinations of \( p_\text{FN} \) and \( p_\text{FP} \) on a small portion of the nuScenes dataset.
We evaluate rendered depths and compare them to the ground truth \ac{LIDAR} measurements as described in \cref{sec:evaluation}.
We then choose the best values for \( p_\text{FN} \) and \( p_\text{FP} \) according to~\cref{fig:heatmap}.
Note that we run the grid search on a small subset of the nuScenes dataset, but use the same hyperparameters for the evaluation on the Waymo dataset.
This highlights the method's applicability for different sensor setups.

\subsection{Training}\label{subsec:training}

We use our generated \ac{GT} data to train existing occupancy prediction methods, with only minor modifications of the original approaches.
Throughout the experiments we keep all model specific parameters fixed and only change the last layer.
All models are trained and evaluated using the nuScenes dataset.
We choose Occ3D's occupancy \acs{GT}~\cite{tian2023occ3d} as reference and compare it to training with our occupancy \acs{GT}.
In the following, we introduce our baseline, as well as two modes of supervision to make use of our training data.

\paragraph{Baseline.}
As detailed in the related work,
BEVDet4D-Occ with ResNet50 as backbone by Huang et al.~\cite{huang2021bevdet, huang2022bevdet4d} serves as our main baseline.
We start from a checkpoint pre-trained on the data of the \textit{CVPR 2023 Occupancy Prediction Challenge}~\cite{tong2023scene, tian2023occ3d}.
Since the original model is trained for semantic occupancy prediction, we fine-tune it for the task of occupancy prediction without semantics.
Therefore, we map all occupied classes to a single binary occupancy target.
We use the sigmoid activation function and the \acl{BCE} as a loss function.
Similar to the pre-training procedure, we use the camera visibility mask provided in the challenge to only supervised voxels, which are labeled as visible.
We use a small learning rate of \num{1e-5} and fine-tune for three epochs.
All subsequent experiments follow the same fine-tuning procedure.
We only modify the last layer, the loss function and the \acs{GT} per-voxel supervision.

\paragraph{\Acf{BCE}.}
The first way to use our training data is to model a binary decision between the occupied belief \(\func{m_i}`*(\text{o})\) and free belief \(\func{m_i}`*(\text{f})\), which is weighted with the uncertainty term \((1 - \func{m}`*(\Omega))\).
We calculate the target probability as:
\begin{equation}
	p^\text{tgt}_i = \frac{\func{m_i}`*(\text{o})}{\func{m_i}`*(\text{o}) + \func{m_i}`*(\text{f})}.
\end{equation}
The total loss is the weighted sum of \acs{BCE} values over all voxels \(i\):
\begin{equation}
	\mathcal{L} = \sum_i (1 - \func{m_i}`*(\Omega)) \mathcal{L}_\text{BCE}(p^\text{tgt}_i, p^\text{pred}_i).
\end{equation}
This setup exactly matches our baseline, except that we exchange the visibility mask with our uncertainty weighting and use our per-voxel target probabilities.

\paragraph{\Acf{CE}.}
Alternatively, we can directly estimate the voxel \ac{BBA} by predicting the mass values of its \textit{occupied}, \textit{free} and \textit{uncertain} hypotheses.
Since we model a distribution over the three hypotheses, we choose the softmax activation function, which automatically enforces their sum to be one.
Following the standard multi-class classification setup, we train the network with the \ac{CE} loss and additionally weight each term with an uncertainty term \((1 - \func{m}`*(\Omega))\):
\begin{equation}
	\mathcal{L} = \sum_i (1 - \func{m_i}`*(\Omega)) \cdot \mathcal{L}_\text{CE}(\vec{p}^\text{tgt}_i, \vec{p}^\text{pred}_i).
\end{equation}
Note that \(\vec{p}^\text{tgt}_i\) and \(\vec{p}^\text{pred}_i\) are now distributions over \textit{occupied}, \textit{free} and \textit{uncertain} hypotheses rather than single values.
A quick overview of the different training modes and supervision signals is given in \cref{tab:models}.

\begin{table*}[t]
	\vspace*{1cm}
	\centering
	\begin{tabular}{l r r r r}
		\toprule
		Name                     & \Acl{GT}                   & Predicted classes                      & Loss     & Loss weighting             \\
		\midrule
		Baseline w/o fine-tuning & Occ3D~\cite{tian2023occ3d} & 16 occupied + 1 free class             & \acs{CE}  & Camera visibility          \\
		Baseline                 & Occ3D~\cite{tian2023occ3d} & \( \{ \text{f}, \text{o}\} \)          & \acs{BCE} & Camera visibility          \\
		Ours \acs{BCE}                 & Our \Ac{BBA}               & \( \{ \text{f}, \text{o}\} \)          & \acs{BCE} & \(1 - \text{m}_i(\Omega)\) \\
		Ours \acs{CE}                  & Our \Ac{BBA}               & \( \{ \text{f}, \text{o}, \Omega \} \) & \acs{CE}  & \(1 - \text{m}_i(\Omega)\) \\

		\bottomrule
	\end{tabular}
	\caption{%
		\textbf{Overview of Parameters Used for Training Occupancy Prediction.}
		We choose BEVDet4D R50~\cite{huang2021bevdet} as our baseline.
		We fine-tune the baseline for the task of binary occupancy prediction using the occupancy maps from Occ3D~\cite{tian2023occ3d}.
		Therefore, we replace the last layer to predict a single occupancy value and apply \acl{BCE} as loss function.
		``Ours BCE'' is fine-tuned the exact same way, we only change the supervision signal by using our data.
		``Ours CE'' additionally predicts the uncertainty mass of our \ac{BBA} and is therefore trained with the \acl{CE} loss.
		For all models, we only change the last layer and the supervision signal, but keep all other parameters the same.
	}%
	\label{tab:models}
\end{table*}

\begin{table*}[t]
	\centering
	\begin{tabular}{l r r r r r r r r}
		\toprule
		Method                                                & \acs{MAE} \lowerbetter & \acs{RMSE} \lowerbetter & \acs{RMSE}$_{log}$ \lowerbetter    & \( \delta < 1.25 \) \higherbetter & \( \delta < 1.25^2 \) \higherbetter & \( \delta < 1.25^3 \) \higherbetter \\
		                                                      & in \unit{\metre} & in \unit{\metre}  & in \( \log{\unit{\metre}} \) & in \%                             & in \%                               & in \%                               \\
		\midrule
		Baseline without fine-tuning \cite{huang2022bevdet4d} & 1.93             & 4.30              & 0.31                         & 74.7                              & 91.5                                & 95.0                                \\
		Baseline \cite{huang2022bevdet4d}                     & 1.89             & 4.26              & 0.30                         & 76.4                              & 92.3                                & 95.4                                \\
		Ours \acs{BCE}                                              & \ranktwo{1.44}   & \ranktwo{3.60}    & \ranktwo{0.25}               & \ranktwo{88.0}                    & \ranktwo{94.0}                      & \ranktwo{96.7}                      \\
		Ours \acs{CE}                                               & \rankone{1.42}   & \rankone{3.47}    & \rankone{0.24}               & \rankone{88.5}                    & \rankone{94.5}                      & \rankone{97.0}                      \\
		\bottomrule
	\end{tabular}
	\caption{%
		\textbf{Comparison of Depth Errors Between Occupancy Predictions and \acs{LIDAR} Measurements.}
		We report errors on rendered depth values from occupancy grids vs.\ raw \ac{LIDAR} measurements.
		We use depth evaluation metrics to quantify the degree of agreement between occupancy map and raw \ac{LIDAR} measurement, which are calculated on the validation set of nuScenes dataset~\cite{caesar2019nuscenes}.
		We only consider rays which are fully enclosed in the volume specified in \cref{tab:params}.
	}
	\label{tab:method_vs_lidar}
\end{table*}

\subsection{Occupancy Prediction}

We compare the predicted occupancy maps with the \ac{LIDAR} measurements using the evaluation protocol described in~\cref*{sec:evaluation}.
We report quantitative results in \cref{tab:method_vs_lidar}.
Models trained with our data outperform the baselines across all metrics with respect to the agreement with the \ac{LIDAR} scans.
We notice that fine-tuning for binary occupancy estimation using the occupancy maps from~\cite{tian2023occ3d} only yields small improvements, whereas fine-tuning with our occupancy maps improves performance significantly.
We attribute this performance gain to our improved occupancy maps, since we only changed the supervision signal.
The prediction of \textit{occupied}, \textit{free} and \textit{uncertain} hypotheses instead of a binary decision improves performance even further, but the improvement is small compared to the improvement due to better \acs{GT} data.
Please see the supplementary material for an additional evaluation of our uncertainty estimates for the \acs{GT} data and model predictions.

\section{Conclusion}\label{sec:conclusion}

In this paper we presented a novel approach for \acl{GT} occupancy data generation from \ac{LIDAR} measurements.
Based on the evidence theory, our approach maps \ac{LIDAR} point clouds onto discrete 3D grids encoding belief masses of each voxel belonging to one of three possible states: \textit{occupied}, \textit{free} or \textit{uncertain}.
We compare our generated \acl{GT} to those of popular other methods, and demonstrated that our data much better agrees with the \ac{LIDAR} measurements than that from existing preprocessing strategies.
We also showed that occupancy estimation neural networks trained on our data yield superior prediction results without any  modifications on the model architecture.
An important future research direction includes the incorporation of semantic information into our preprocessing and training pipelines.
\newpage

{\small
	\bibliographystyle{ieeenat_fullname}
	\bibliography{references}
}

\clearpage
\appendix
\section{Video and Images}
\begin{figure}
	\centering
	\begin{minipage}[b]{0.5\textwidth}
		\includegraphics[width=\linewidth]{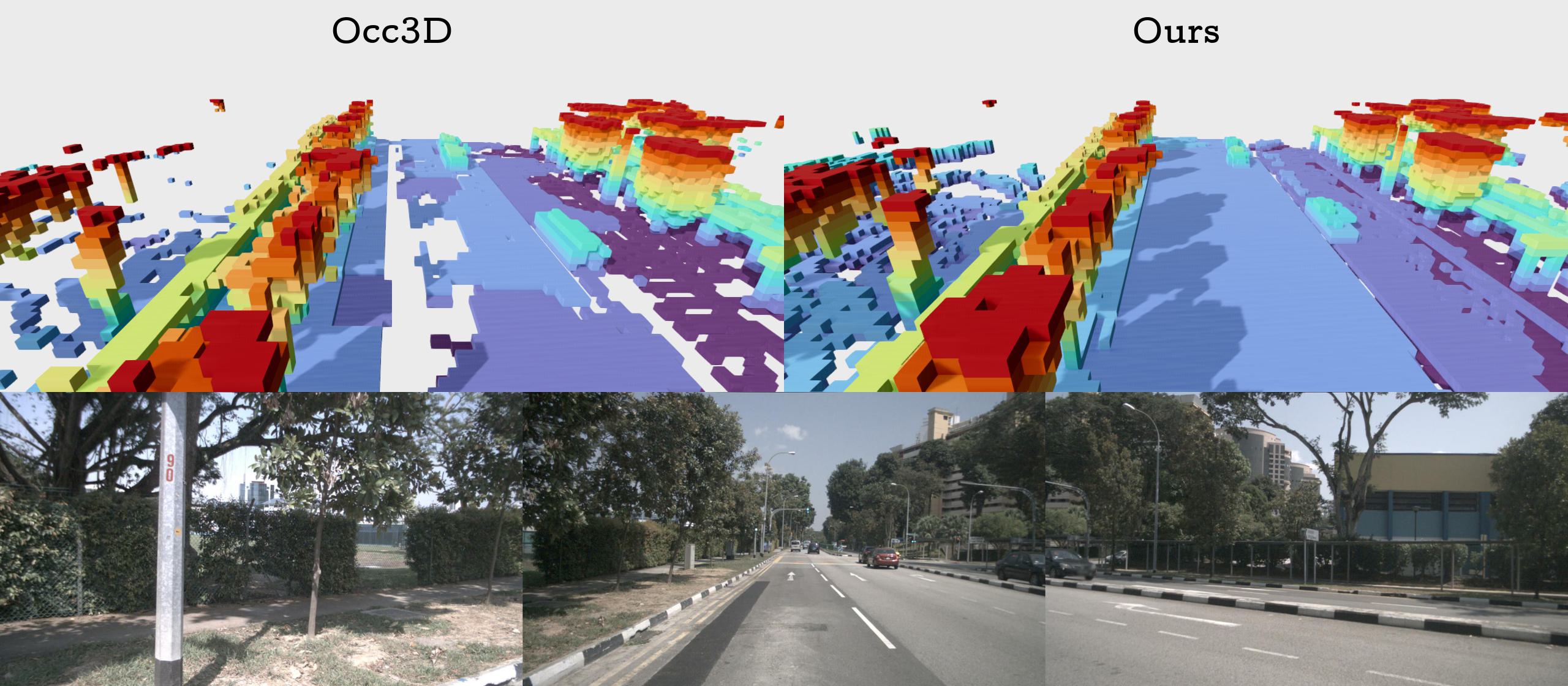}
		\includegraphics[width=\linewidth]{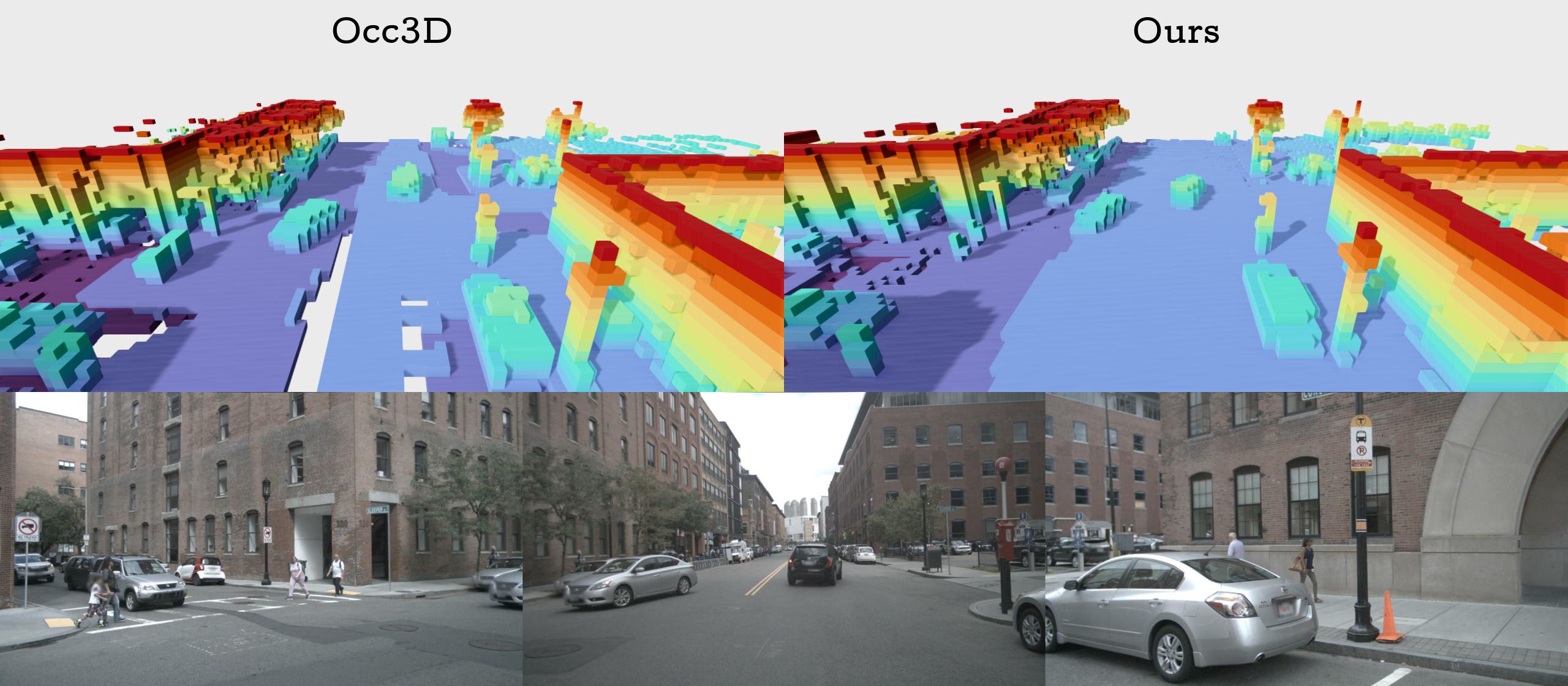}
		\includegraphics[width=\linewidth]{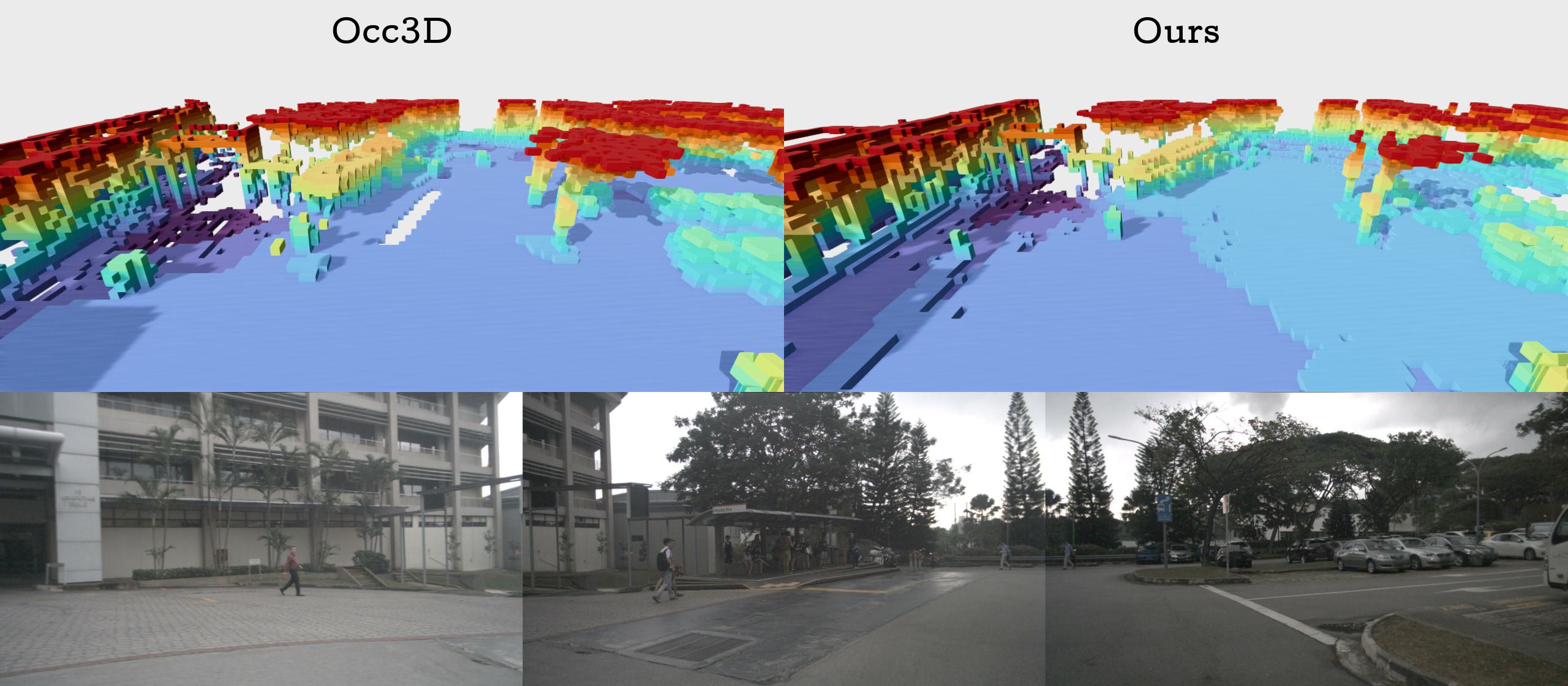}
		\includegraphics[width=\linewidth]{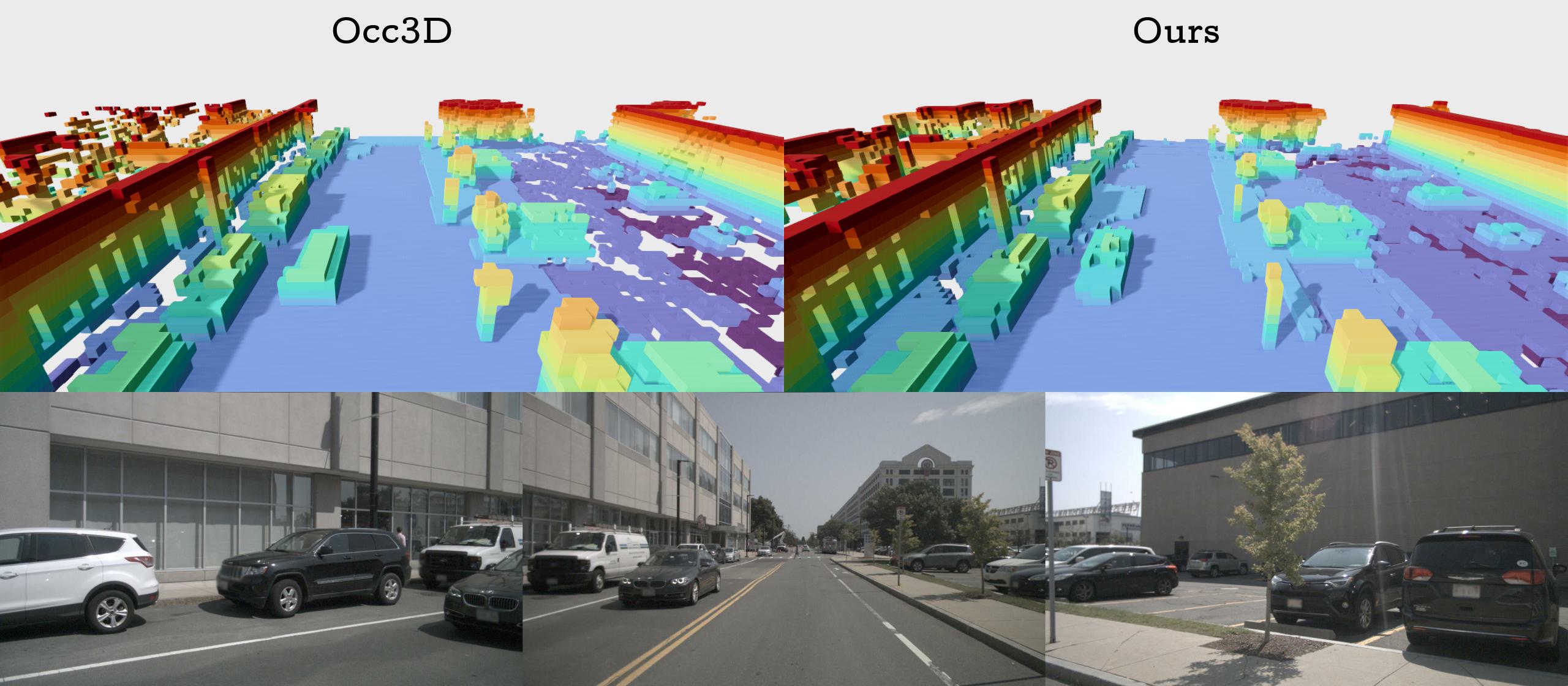}
		\includegraphics[width=\linewidth]{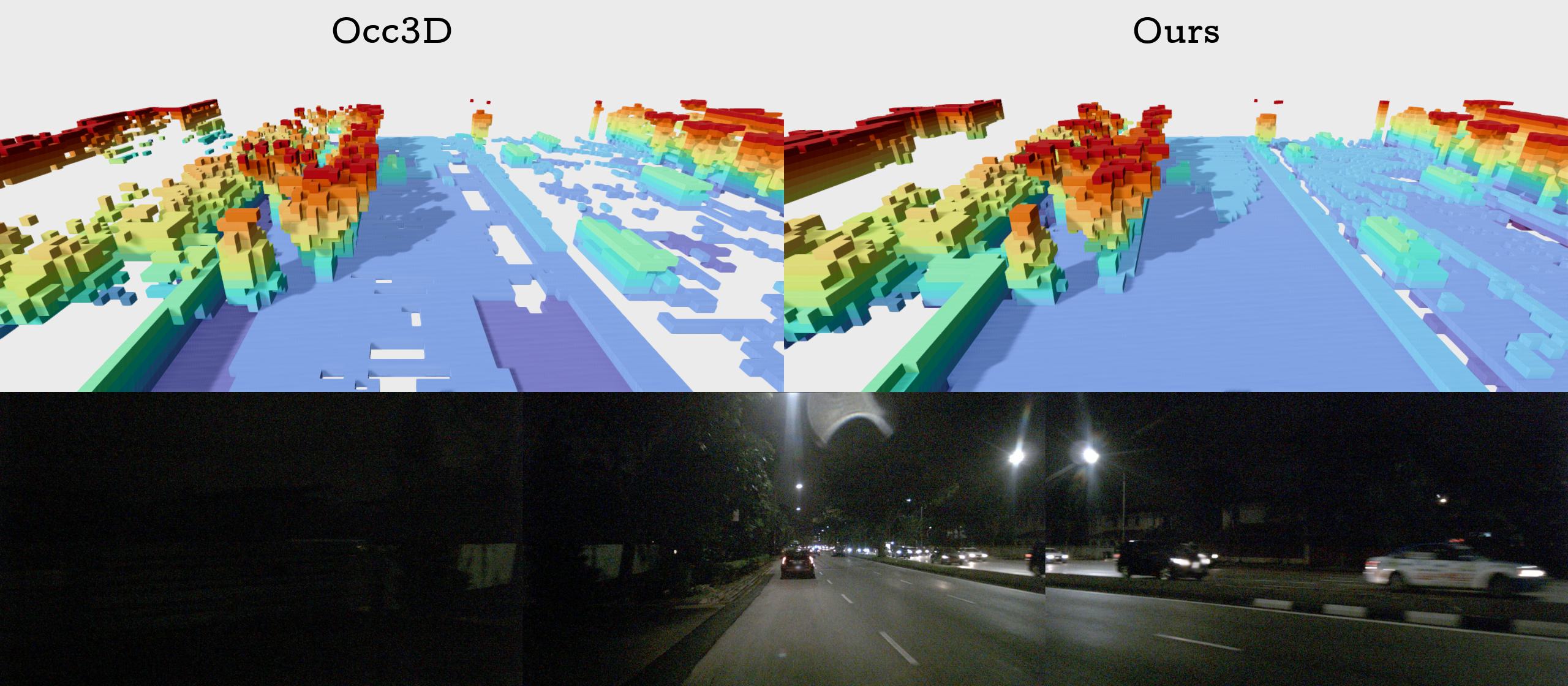}
	\end{minipage}
	\caption{%
		\textbf{Qualitative Comparison on the nuScenes Dataset.}
		We compare the occupancy maps generated by our method with the ones of Occ3D~\cite{tian2023occ3d} on the nuScenes dataset~\cite{caesar2019nuscenes}.
		Our method yields much better street surfaces compared to Occ3D in many scenarios.
		The voxel size of both methods is \qty{0.4}{\metre}.
	}\label{fig:nuscenes_3d}
\end{figure}

\begin{figure}
	\centering
	\begin{minipage}[b]{0.5\textwidth}
		\includegraphics[width=\linewidth]{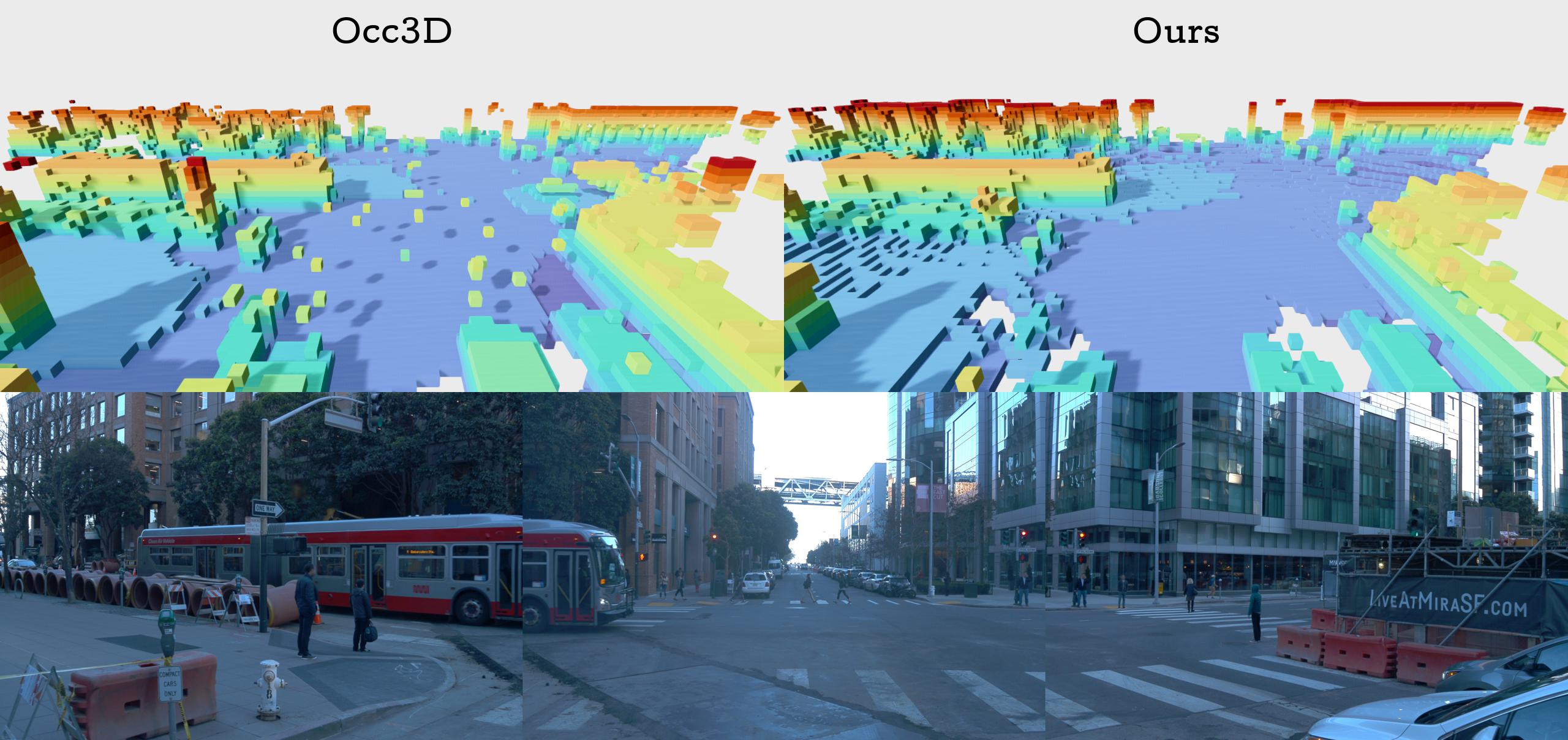}
		\includegraphics[width=\linewidth]{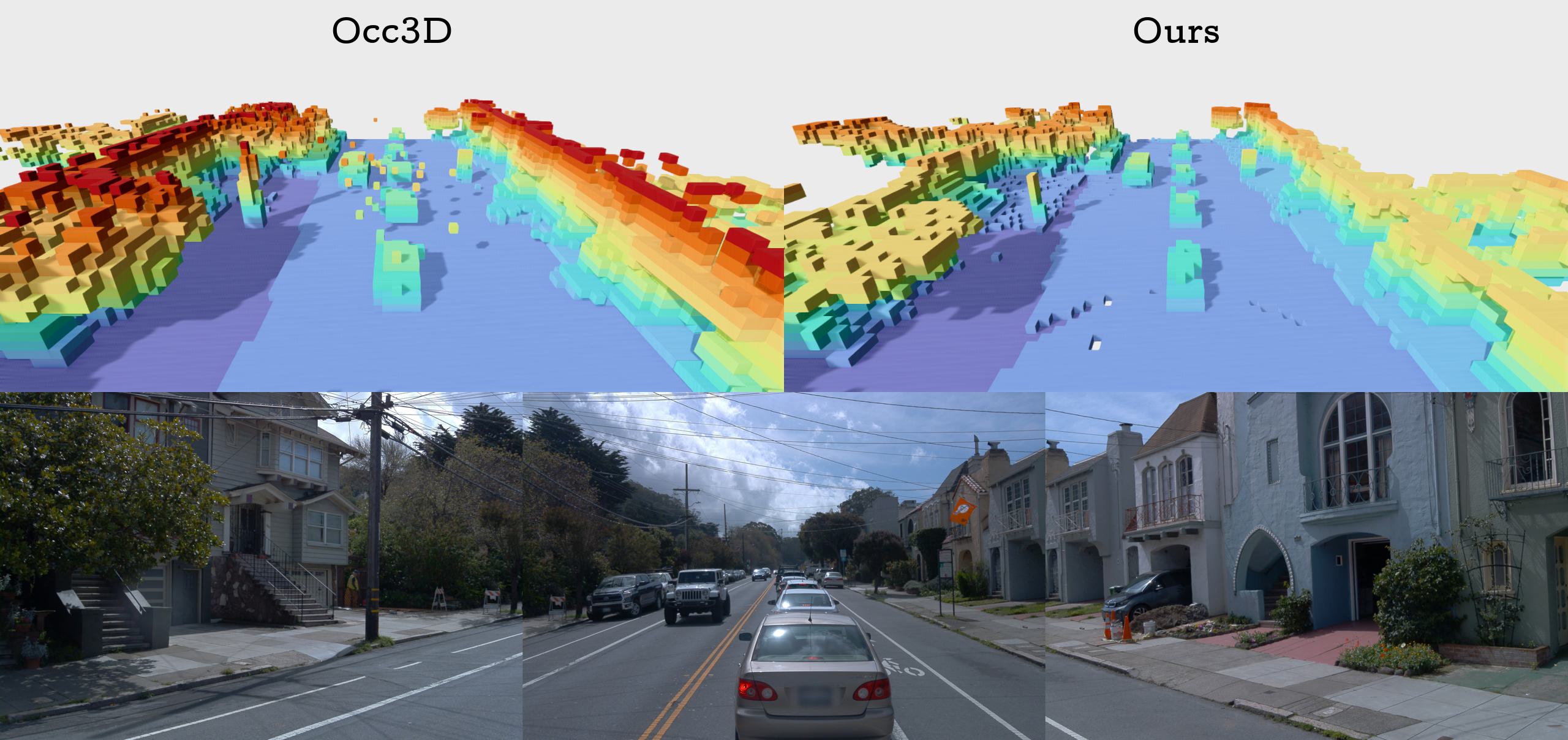}
		\includegraphics[width=\linewidth]{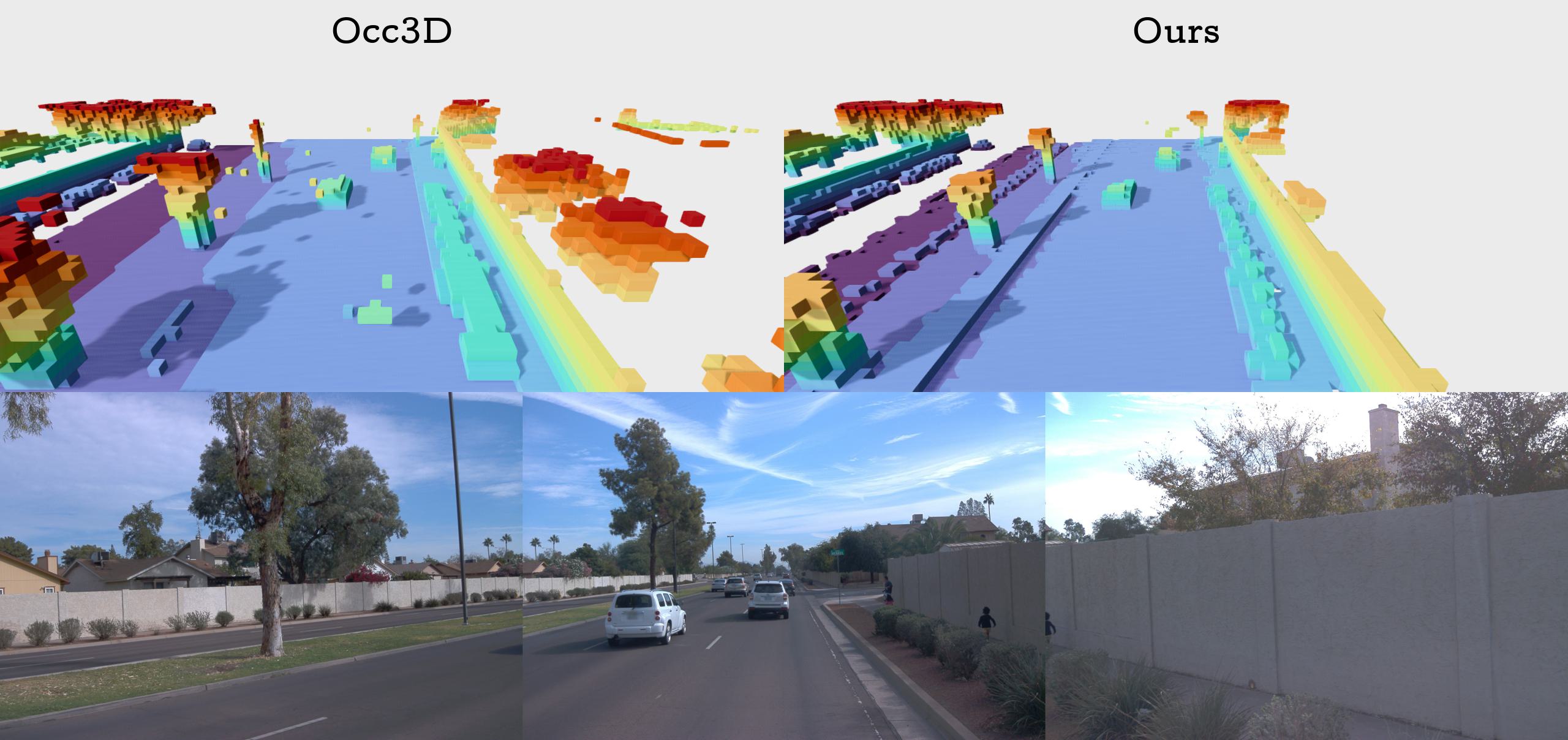}
		\includegraphics[width=\linewidth]{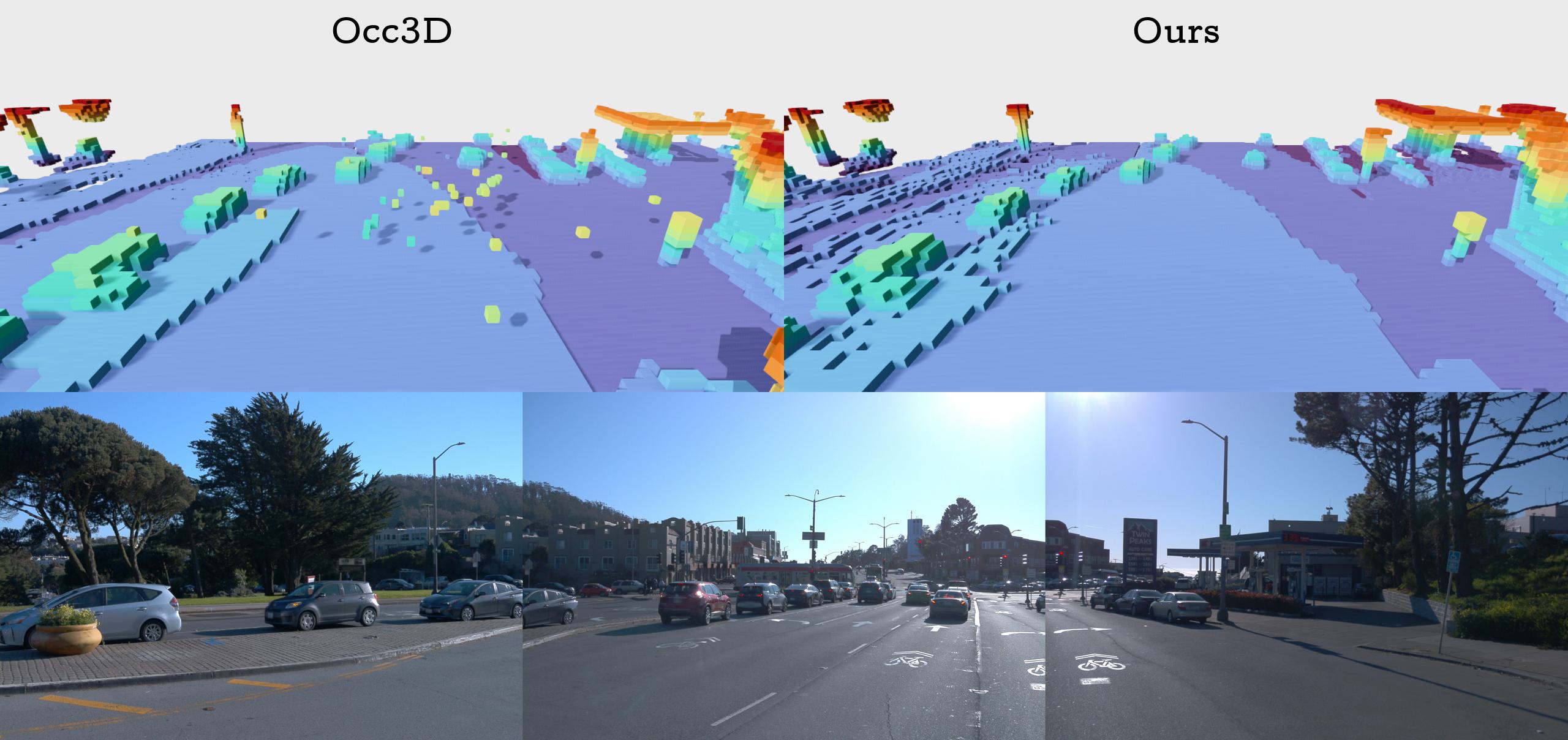}
		\includegraphics[width=\linewidth]{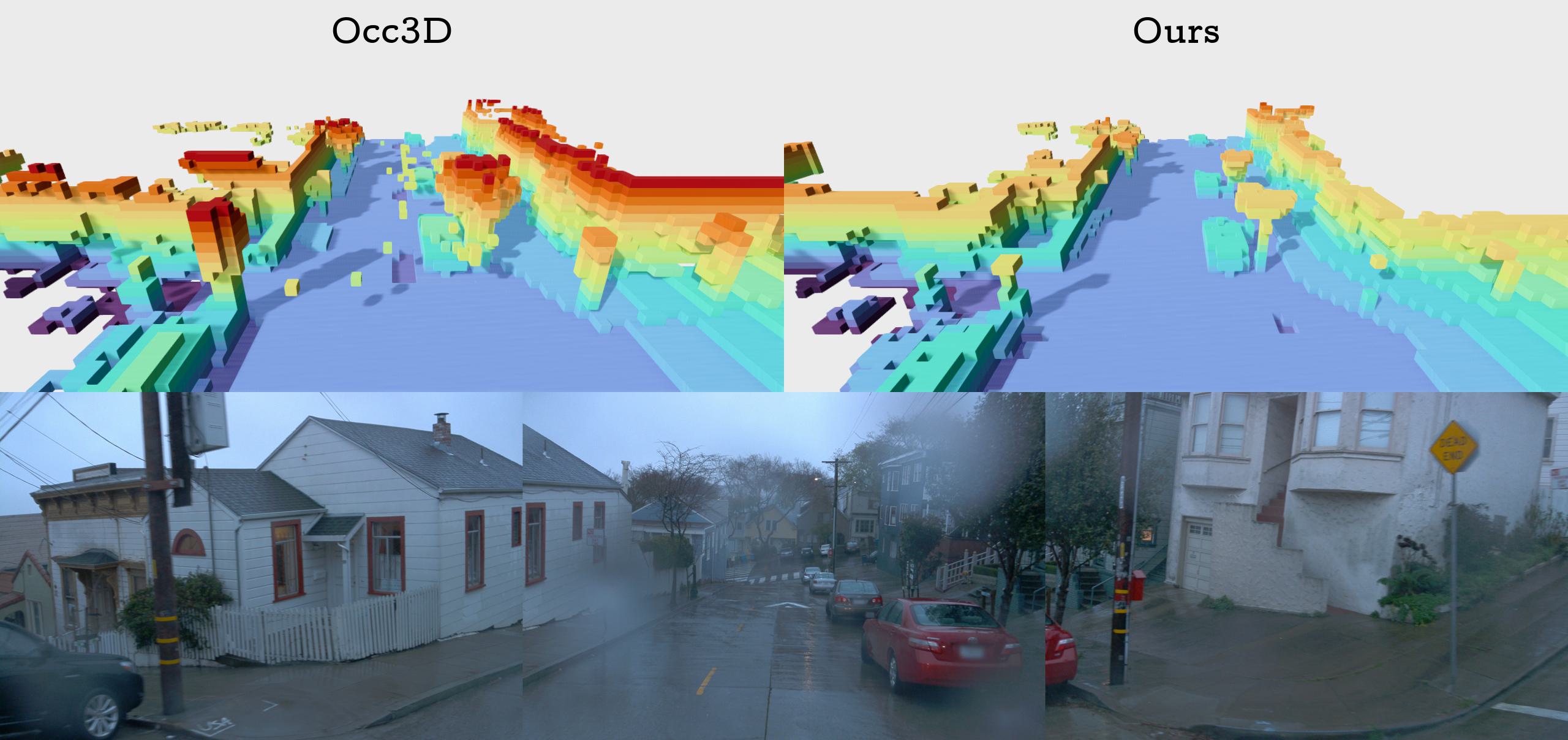}
	\end{minipage}
	\caption{%
		\textbf{Qualitative Comparison on the Waymo Dataset.}
		We compare the occupancy maps generated by our method with the ones of Occ3D~\cite{tian2023occ3d} on the Waymo dataset~\cite{sun2019waymo}.
		Our method yields much less floating artifacts due to the explicit modeling of free space.
        The voxel size of both methods is \qty{0.4}{\metre}.
	}\label{fig:waymo_3d}
\end{figure}

\begin{center}
	\fbox{\begin{minipage}{0.97\columnwidth}
			Please find the video attached with of our rendered depths and examples comparing our approach to the baselines.
			\cref{fig:nuscenes_3d} and \cref{fig:waymo_3d}  illustrates the problem of street surface estimation and floating artifacts on nuScenes and Waymo, respectively.
		\end{minipage}}
\end{center}

\section{Uncertainty Estimation}

Please note that directly evaluating the uncertainty of the beliefs for each voxel is not possible, as no \acl{GT} is available.

To this end, we derive an uncertainty for the rendered depth instead, considering all the voxels through which a ray passes.
Specifically, we use the \ac{BBA} to estimate an upper and lower bound of the depth and take the difference as an uncertainty.
Please recall that in section 4.1., we compared occupied and free belief to decide if a voxel \(i\) is occupied or not.
This is equivalent to distributing the uncertainty $\func{m_i}`*(\Omega)$ equally to the occupied and free hypothesis:
\begin{align}
	o_i & =
	\begin{dcases*}
		1 & if \(\func{m_i}`*(\text{o}) > \func{m_i}`*(\text{f})\) \\
		0 & else
	\end{dcases*} \\
	    & =
	\begin{dcases*}
		1 & if \(\func{m_i}`*(\text{o}) + \frac{1}{2}\func{m_i}`*(\Omega) > \func{m_i}`*(\text{f})+ \frac{1}{2}\func{m_i}`*(\Omega) \) \\
		0 & else
	\end{dcases*}.
\end{align}
The depth $d^\text{est}_j$ is then normally determined by finding the first occupied voxel $o_i$ along the ray.

\paragraph{Minimal and Maximal Ray Length.}
Since the uncertainty mass $\func{m_i}`*(\Omega)$ is compatible with all hypotheses, other assignments of $\func{m_i}`*(\Omega)$ are also valid.
Therefore, we estimate the lower and upper bound of the ray length by distributing the uncertainty mass to only one of both hypotheses.
If we assign all uncertainty mass $\func{m_i}`*(\Omega)$ to the occupied hypothesis $\func{m_i}`*(\text{o})$, we obtain more occupied voxels:
\begin{align}
	o^\text{occ}_i & =
	\begin{dcases*}
		1 & if \(\func{m_i}`*(\text{o}) + \func{m_i}`*(\Omega) > \func{m_i}`*(\text{f})\) \\
		0 & else
	\end{dcases*}.
\end{align}
This leads to the the minimal ray length $d^\text{min}_j$.
Contrary to this, assigning all uncertainty to the free hypothesis:
\begin{align}
	o^\text{free}_i & =
	\begin{dcases*}
		1 & if \(\func{m_i}`*(\text{o}) > \func{m_i}`*(\text{f}) + \func{m_i}`*(\Omega)\) \\
		0 & else
	\end{dcases*}
\end{align}
will lead to less occupied voxels, and we obtain the maximum ray length $d^\text{max}_j$.
Overall, we obtain the lower bound of the ray length $d^\text{min}_j$, the upper bound of the ray length $d^\text{max}_j$, and the already computed estimated ray length $d^\text{est}_j$.

\begin{figure*}
    \centering
    \includegraphics[width=\linewidth]{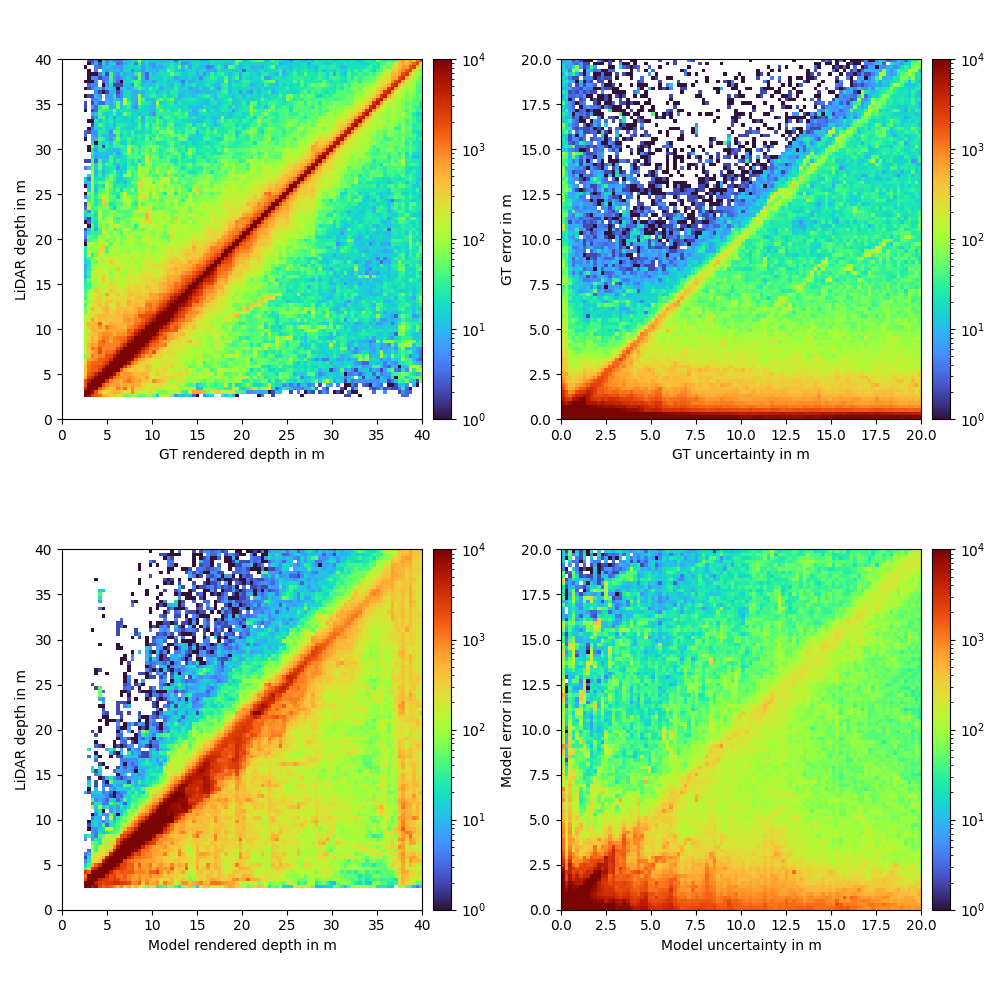}
    \caption{\textbf{Uncertainty Estimates.}
        We compare rendered depths and corresponding uncertainties of our occupancy ground-truth data with our model predictions.
        The top row contains data from our ground-truth occupancy, while the bottom row contains data from model predictions.
        We plot rendered depths $d^\text{est}_j$ against the \acs{LIDAR} measurements on the left side.
        Therefore, we create a heatmap containing the absolute frequency of $(d^\text{est}_j, d^\text{lidar}_j)$ pairs.
        Perfect predictions lie on the diagonal such that $d^\text{est}_j = d^\text{lidar}_j$.        
        The right column shows the estimated uncertainty $d^\text{uncert}_j$ on the $x$-axis and the estimation error $d^\text{error}_j = |d^\text{est}_j - d^\text{lidar}_j|$ on the $y$-axis.
        We follow the same procedure to create the heatmap of $(d^\text{uncert}_j, d^\text{error}_j)$ pairs.     
        As visible by the strong diagonals, our method provides meaningful uncertainty estimates with a slight tendency to overestimate them. }
    \label{fig:uncertainty}
\end{figure*}

\paragraph{Rendered Depth Uncertainty.}
Given the bounds, we define the uncertainty per ray as maximal deviation from the estimation:
\begin{align}
	d^\text{uncert}_j & =
	\max{\left(|d^\text{max}_j - d^\text{est}_j|, |d^\text{est}_j - d^\text{min}_j|\right)}.
\end{align}
Note, that all ray lengths being equal corresponds to an uncertainty of zero.
We compare the obtained uncertainty per ray $d^\text{uncert}_j$ with the error to the \acs{LIDAR} measurement
\begin{align}
	d^\text{error}_j & =
	|d^\text{est}_j - d^\text{lidar}_j| \,.
\end{align}
Please find the results illustrated in \cref{fig:uncertainty}.

\section{Multi-Frame Temporal Aggregation}
In \cref{tbl:temporal_aggregation} we show the impact of changing the number of frames for temporal aggregation.
We observe that more frames for temporal aggregation increase the performance at the cost of computation time.

\begin{table}[ht]
	\centering
	\small
	\setlength{\tabcolsep}{4pt} %
	\begin{tabular}{l r r r r r r r r}
		\toprule
		\#frames & \acs{MAE}                            & \acs{RMSE}                           & \( \delta\!<\!1.25 \) & \( \delta\!<\!1.25^2 \) & \( \delta\!<\!1.25^3 \) \\
		         & \lowerbetter~ in \unit{\metre} & \lowerbetter~ in \unit{\metre} & \higherbetter~ in \%  & \higherbetter~ in \%    & \higherbetter~ in \%    \\
		\midrule
		5        & 0.99                           & 3.09                           & 90.0                  & 94.8                    & 97.2                    \\
		10       & 0.94                           & 2.99                           & 91.8                  & 95.6                    & 97.4                    \\
		50       & 0.92                           & 2.96                           & 92.9                  & 96.2                    & 97.7                    \\
		100      & 0.92                           & 2.95                           & 93.0                  & 96.3                    & 97.7                    \\
		\bottomrule
	\end{tabular}
	\caption{%
		\textbf{Impact of Number of Frames Used for Temporal Aggregation.}
		We evaluate the generated \acs{GT} for a varying number of frames used for temporal aggregation on the nuScenes mini dataset with a voxel size $\qty{0.2}{\metre}$.
	}
	\label{tbl:temporal_aggregation}
	\vspace{-0.8em}
\end{table}

\section{Uncertainty Loss Weighting}
To further analyze the influence of uncertainty weighting, we trained  the models without it and report the results in \cref{tbl:loss_weighting}.
Uncertainty weighting is important for the final performance.
\begin{table}[ht]
	\centering
	\small
	\begin{adjustbox}{width=\columnwidth}
		\begin{tabular}{l r r r r r}
			\toprule
			Loss      & \acs{MAE}                      & \acs{RMSE}                     & \( \delta < 1.25 \)  & \( \delta < 1.25^2 \) & \( \delta < 1.25^3 \) \\
			          & \lowerbetter~ in \unit{\metre} & \lowerbetter~ in \unit{\metre} & \higherbetter~ in \% & \higherbetter~ in \%  & \higherbetter~ in \%  \\
			\midrule
			\acs{BCE} & 1.44/1.70                      & 3.60/4.04                      & 88.0/82.7            & 94.0/91.1             & 96.7/94.6             \\
			\acs{CE}  & 1.42/1.55                      & 3.47/4.10                      & 88.5/88.2            & 94.5/94.0             & 97.0/96.5             \\
			\bottomrule
		\end{tabular}
	\end{adjustbox}
	\caption{
		\textbf{Uncertainty Loss Weighting.} Training results with/without uncertainty loss weighting.
	}	\label{tbl:loss_weighting}
\end{table}

\end{document}